\documentclass[runningheads]{eccv_2024/llncs}

\usepackage{eccv_2024/eccv}

\usepackage{eccv_2024/eccvabbrv}

\usepackage{graphicx}
\usepackage{booktabs}

\usepackage{tabularx}
\usepackage{graphicx}
\usepackage[dvipsnames]{xcolor}
\usepackage{colortbl}
\usepackage{svg}
\usepackage{bbm}
\usepackage{multirow}
\usepackage{wasysym}
\usepackage{stix}
\usepackage{amsmath,amsfonts,amssymb}

\usepackage{colortbl}
\usepackage{caption}
\usepackage{subcaption}
\usepackage{comment}
\usepackage{stix}
\usepackage[accsupp]{axessibility}  

\usepackage{hyperref}

\usepackage{orcidlink}

\usepackage{tikz}
\usetikzlibrary{shapes}
\usetikzlibrary{plotmarks}
\usetikzlibrary{patterns}
\usepackage{ifthen}

\usepackage{pgfplots}
\pgfplotsset{width=10cm,compat=1.18}
\usepgfplotslibrary{statistics}
\usepgfplotslibrary{colormaps}
\usepgfplotslibrary{fillbetween}
\usetikzlibrary{pgfplots.groupplots}
\pgfplotsset{compat=newest}

\usepackage{pgfplotstable}

\graphicspath{{./images/}} 

\definecolor{dc1}{RGB}{230, 25, 75} 
\definecolor{dc2}{RGB}{60, 180, 75} 
\definecolor{dc3}{RGB}{255, 225, 25} 
\definecolor{dc4}{RGB}{0, 130, 200} 
\definecolor{dc5}{RGB}{245, 130, 48} 
\definecolor{dc6}{RGB}{145, 30, 180} 
\definecolor{dc7}{RGB}{70, 240, 240} 
\definecolor{dc8}{RGB}{240, 50, 230} 
\definecolor{dc9}{RGB}{210, 245, 60} 
\definecolor{dc10}{RGB}{250, 190, 212} 
\definecolor{dc11}{RGB}{0, 128, 128} 
\definecolor{dc12}{RGB}{220, 190, 255} 
\definecolor{dc13}{RGB}{170, 110, 40} 
\definecolor{dc14}{RGB}{255, 250, 200} 
\definecolor{dc15}{RGB}{128, 0, 0} 
\definecolor{dc16}{RGB}{170, 255, 195} 
\definecolor{dc17}{RGB}{128, 128, 0} 
\definecolor{dc18}{RGB}{255, 215, 180} 
\definecolor{dc19}{RGB}{0, 0, 128} 
\definecolor{dc20}{RGB}{128, 128, 128} 
\definecolor{dc21}{RGB}{255, 255, 255} 
\definecolor{dc22}{RGB}{0, 0, 0} 

\definecolor{mylightgray}{gray}{0.9}
\definecolor{tgraspable}{RGB}{255, 0, 0}
\definecolor{tcut}{RGB}{0, 0, 255}
\definecolor{tscoop}{RGB}{255, 255, 0}
\definecolor{tcontain}{RGB}{0, 255, 0}
\definecolor{tpound}{RGB}{128, 80, 215}
\definecolor{tsupport}{RGB}{0, 255, 255}
\definecolor{twgrasp}{RGB}{196, 164, 132}
\definecolor{tarm}{RGB}{255, 255, 255}

\begin{document}

\title{Segmenting Object Affordances:\\Reproducibility and Sensitivity to Scale} 

\titlerunning{Segmenting object affordances}

\author{Tommaso Apicella\inst{1,2}\orcidlink{0000-0001-9001-5641} \and
Alessio Xompero\inst{2}\orcidlink{0000-0002-8227-8529} \and
Paolo Gastaldo\inst{1}\orcidlink{0000-0002-5748-3942
}\and \\
Andrea Cavallaro\inst{3,4}\orcidlink{0000-0001-5086-7858}}

\authorrunning{T.~Apicella et al.}

\institute{University of Genoa, Italy \and
Queen Mary University of London, United Kingdom
 \and
Idiap Research Institute, Switzerland \and 
\'{E}cole Polytechnique F\'{e}d\'{e}rale de Lausanne, Switzerland\\
\email{\{t.apicella, a.xompero\}@qmul.ac.uk} \\ 
\email{paolo.gastaldo@unige.it}\\  
\email{a.cavallaro@idiap.ch}}

\maketitle

\begin{abstract}
Visual affordance segmentation identifies image regions of an object an agent can interact with. Existing methods re-use and adapt learning-based architectures for semantic segmentation to the affordance segmentation task and evaluate on small-size datasets. However, experimental setups are often not reproducible, thus leading to unfair and inconsistent comparisons. In this work, we benchmark these methods under a reproducible setup on two single objects scenarios, tabletop without occlusions and hand-held containers, to facilitate future comparisons. We include a version of a recent architecture, Mask2Former, re-trained for affordance segmentation and show that this model is the best-performing on most testing sets of both scenarios. Our analysis shows that models are not robust to scale variations when object resolutions differ from those in the training set.
\end{abstract}

\section{Introduction}~\label{sec:intro}
Affordances are the potential actions that objects of interest present in the environment offer to an agent (e.g., a person or a robot)~\cite{gibson1966senses}. This definition implies that an object can suggest two main interactions to accomplish an action: one between the agent and the object (agent-object) and another between the object and the environment (object-object)~\cite{osiurak2017affordance}. For example, to fill a cup (action), a person needs to first grasp the cup (agent-object) and then pour the filling into the cup (object-object). 
Visual affordance segmentation identifies the affordance regions of an object directly from an image~\cite{myers2015affordance, hassanin2021visual}.
In the filling example, the external surface of the cup is a \textit{graspable} region, whereas the internal part can \textit{contain} the filling. Affordance segmentation is important for assistive robotic and human-robot collaboration~\cite{myers2015affordance, nguyen2017object, yin2022object, yin2022new, pang2021towards, rosenberger2020object, sanchez2020benchmark, yang2021reactive}, action recognition~\cite{kjellstrom2011visual} and anticipation~\cite{thakur2024leveraging}, and augmented and virtual reality~\cite{hassan2021populating}. 

However, variations in the physical properties and appearance of the objects of interest as well as  occlusions make the task challenging. Objects of the same type, such as drinking cups or knives, can vary in shape, color, and material~\cite{myers2015affordance, nguyen2017object}, or different object types can be designed for different actions, but their perception in an image can result in visually similar regions. For example, the surface of a trowel to \textit{scoop} with and the surface of a turner to \textit{support} with. 
Occlusions can occur when other objects are present in front of the object of interest and with respect to the camera observing the object, or when the object is hand-held by a person (hand-occlusion)~\cite{apicella2023affordance, xompero2022corsmal}. These occlusions result in the object of interest only partially observed in the image and increase the difficulty of accurately identifying the visible affordance regions. Moreover, the recognition and segmentation of these regions is further affected by the perspective of the object in more challenging poses when hand-held by a person~\cite{apicella2023affordance, xompero2022corsmal}.

To tackle these challenges, most of the existing methods independently assign an affordance class to each pixel and group regions with pixels belonging to the same class~\cite{hussain2020fpha, nguyen2016detecting, nguyen2017object}. However, when the object is not close to the camera or not easily distinguishable from the background, these methods can misclassify areas outside of the object region. Other methods~\cite{gu2021visual, yin2022new, zhang2022multi, zhao2020object} use attention mechanisms to select relevant pixel positions or relevant channels within low-resolution feature maps extracted from the image, while avoiding the high computational cost at higher resolutions. However, the coarse spatial information of low-resolution feature maps can affect the affordance segmentation, degrading the details of object regions. In the attention mechanism, latent vectors can replace feature maps as query, enabling higher resolutions through a more efficient computation~\cite{jaegle2021perceiver}. Latent vectors, also referred to as object queries, are compact representations independent from the input and can be learned during training time. Latent vectors can be combined with the feature maps to predict specific object classes location~\cite{carion2020end} or segmentation masks~\cite{strudel2021segmenter, cheng2021per,cheng2022masked}. Despite these advantages, no existing work modelled latent vectors for visual affordance segmentation. Moreover, a fair comparison with previous works is not always possible due to unavailable implementations or weights of the model, and performance is compared quantitatively in spite of changes in the training setup~\cite{do2018affordancenet, gu2021visual, nguyen2016detecting}. Previous works evaluated models using the benchmarks and comparing results, without considering how model could perform in a real scenario, where the camera might be closer or farther from the object of interest. In case of robotic experiments, the camera pose was fixed in a way that could resemble the distance of the object observed during training~\cite{chen2023survey, do2018affordancenet, nguyen2016detecting, nguyen2017object, yin2022new, yin2022object}.

In this paper, we provide an up-to-date benchmark by re-implementing and re-training existing methods under the same framework and setup, along with a recent architecture for semantic segmentation. Through our benchmark, the community can further investigate the affordance segmentation problem in a fair and reproducible way\footnote[1]{\scriptsize{Code and trained models are available at \url{https://apicis.github.io/aff-seg}}}. We show that models trained under the same setup have lower performance than their performance in the respective papers in the scenario of a single object on a tabletop. We assess  generalisation performance of the models by varying the object scale and using the object occupancy i.e., the ratio of pixels belonging to the object mask and the total number of pixels, as proxy for the scale. To complement the variety of architectures that tackle the affordance segmentation problem, we retrain Mask2Former\cite{cheng2022masked}, one of the most recent methods using latent vectors for segmentation. Our experiments show that Mask2Former outperforms other methods on the dataset with single object on a tabletop, and on most datasets with hand-occluded objects.

\section{Related work}
\label{sec:related_works}

Methods for visual affordance segmentation use semantic and instance segmentation to predict affordance regions on the objects~\cite{badrinarayanan2015segnet,chen2017deeplab, fu2019dual,he2017mask,ronneberger2015u,wu2019fastfcn,zhao2017pyramid}. 
Table~\ref{tab:relatedworks} summarises the characteristics of existing models for affordance segmentation discussed in this section. The training and evaluation of models is performed with different datasets (see Table~\ref{tab:affdatasets}), based on the chosen object instances and the affordance classes. Benchmarks are testing sets of datasets, held out during training.   

\subsection{Methods for affordance segmentation}

\begin{table}[t!]
    \centering
    \scriptsize
    \caption{Characteristics and comparison of visual affordance segmentation models. Note that we report the best-performing backbone for each model, and we do not consider additional parts of the pipelines, such as a separate object detector.
    \vspace{-7pt}
    }
    \begin{tabular}{r r r c c cccc cc ccc c c}
    \toprule
    \textbf{Model} & \textbf{Source architecture} & \textbf{Backbone} & \textbf{FPN} & \textbf{IF} & \multicolumn{4}{c}{\textbf{Attention}} & \multicolumn{2}{c}{\textbf{Aff.}} & \multicolumn{3}{c}{\textbf{Object}} & \textbf{CRF} & \textbf{FS}\\
    \cmidrule(lr){6-9} \cmidrule(lr){10-11} \cmidrule(lr){12-14}
    & & & & & Sp & Ch & Sa & Mc & C & E & C & S & L & & \\
    \midrule
    AffordanceNet~\cite{do2018affordancenet} & Mask R-CNN~\cite{he2017mask} & VGG-16~\cite{simonyan2014very} & $\circ$ & $\circ$ & $\circ$ & $\circ$ & $\circ$ & $\circ$ & $\circ$ & $\circ$ & $\bullet$ & $\circ$ & $\bullet$ & $\circ$ & $\bullet$\\
    B-Mask R-CNN~\cite{mur2023bayesian} & Mask R-CNN~\cite{morrison2019uncertainty, minh2020learning} & RNX-101~\cite{xie2017aggregated} & $\bullet$ & $\circ$ & $\circ$ & $\circ$ & $\circ$ & $\circ$ & $\circ$ & $\circ$ & $\bullet$ & $\circ$ & $\bullet$ & $\circ$ & $\bullet$ \\
    4C-RPN-5C~\cite{minh2020learning} & AffordanceNet~\cite{do2018affordancenet} & SE-RNX-101~\cite{hu2018squeeze} & $\circ$ & $\circ$ & $\circ$ & $\circ$ & $\circ$ & $\circ$ & $\circ$ & $\circ$ & $\bullet$ & $\circ$ & $\bullet$ & $\circ$ & $\bullet$ \\
    A-Mask R-CNN~\cite{caselles2021standard} & AffordanceNet~\cite{do2018affordancenet} & RN-50~\cite{he2016deep} & $\bullet$ & $\circ$ & $\circ$  & $\circ$ & $\circ$ & $\circ$ & $\circ$ & $\circ$ & $\bullet$ & $\circ$ & $\bullet$ & $\circ$ & $\bullet$ \\
    BPN~\cite{yin2022object} & AffordanceNet~\cite{do2018affordancenet} & RN-50~\cite{he2016deep} & $\bullet$ & $\circ$ & $\bullet$ & $\bullet$ & $\circ$ & $\circ$ & $\circ$ & $\bullet$ & $\bullet$ & $\circ$ & $\bullet$ & $\circ$ & $\bullet$ \\
    SEANet~\cite{yin2022new} & DFF~\cite{hu2019dynamic} & RN-50~\cite{he2016deep} & $\circ$ & $\bullet$ & $\bullet$ & $\bullet$ & $\circ$ & $\circ$ & $\circ$ & $\bullet$ & $\circ$ & $\circ$ & $\circ$ & $\circ$ & $\bullet$\\
    BB-CNN~\cite{nguyen2017object} & DeepLab~\cite{chen2017deeplab} & VGG-16~\cite{simonyan2014very} & $\circ$ & $\circ$ & $\circ$ & $\circ$ & $\circ$ & $\circ$ & $\circ$ & $\circ$ & $\circ$ & $\circ$ & $\circ$ & $\bullet$ & $\bullet$ \\
    DeepLab~\cite{sawatzky2017weakly} & Deeplab~\cite{chen2017deeplab} & RN-101~\cite{he2016deep} & $\circ$ & $\circ$ & $\circ$ & $\circ$ & $\circ$ & $\circ$ & $\circ$ & $\circ$ & $\circ$ & $\circ$ & $\circ$ & $\bullet$ & $\circ$ \\
    ADOSMNet~\cite{chen2023adosmnet} & PSPNet~\cite{zhao2017pyramid} & RN-101~\cite{he2016deep} & $\circ$ & $\circ$ & $\circ$ & $\circ$ & $\circ$ & $\circ$ & $\circ$ & $\circ$ & $\circ$ & $\circ$ & $\circ$ & $\circ$ & $\bullet$\\
    CNN~\cite{nguyen2016detecting} & SegNet~\cite{badrinarayanan2015segnet} & VGG-16~\cite{simonyan2014very} & $\circ$ & $\circ$ & $\circ$ & $\circ$ & $\circ$ & $\circ$ & $\circ$ & $\circ$ & $\circ$ & $\circ$ & $\circ$ & $\circ$ & $\bullet$\\
    RN50-F~\cite{hussain2020fpha} & Fast-FCN~\cite{wu2019fastfcn} & RN-50~\cite{he2016deep} & $\circ$ & $\circ$ & $\circ$ & $\circ$ & $\circ$ & $\circ$ & $\circ$ & $\circ$ & $\circ$ & $\circ$ & $\circ$ & $\circ$ & $\bullet$ \\
    ACANet~\cite{apicella2023affordance} & UNet~\cite{ronneberger2015u} & RN-18~\cite{he2016deep} & $\circ$ & $\circ$ & $\circ$ & $\circ$ & $\circ$ & $\circ$ & $\circ$ & $\circ$ & $\circ$ & $\bullet$ & $\circ$ & $\circ$ & $\bullet$ \\
    DRNAtt~\cite{gu2021visual} & DANet~\cite{fu2019dual}  & DRN~\cite{yu2017dilated} & $\circ$ & $\circ$ & $\bullet$ & $\bullet$ & $\circ$ & $\circ$ & $\circ$ & $\circ$ & $\circ$ & $\circ$ & $\circ$ & $\circ$ & $\bullet$ \\
    GSE~\cite{zhang2022multi} & HRNet~\cite{sun2019deep, zhang2017deep} & RNS-101~\cite{zhang2022resnest} & $\circ$ & $\bullet$ & $\circ$ & $\bullet$ & $\circ$ & $\circ$ & $\circ$ & $\circ$ & $\circ$ & $\circ$ & $\circ$ & $\circ$ & $\bullet$ \\
    RANet~\cite{zhao2020object} & EncNet~\cite{zhang2018context} & RN-50~\cite{he2016deep} & $\circ$ & $\circ$ & $\circ$ & $\bullet$ & $\circ$ & $\circ$ & $\bullet$ & $\circ$ & $\bullet$ & $\circ$ & $\circ$ & $\circ$ & $\bullet$ \\
    STRAP~\cite{cui2023strap} & SINN~\cite{nauata2019structured} & RN-50~\cite{he2016deep} & $\circ$ & $\bullet$ & $\circ$ & $\circ$ & $\bullet$ & $\circ$ & $\bullet$ & $\circ$ & $\circ$ & $\circ$ & $\circ$ & $\bullet$ & $\circ$ \\
    M2F-Aff & Mask2Former~\cite{cheng2022masked, cheng2021per} & RN-50~\cite{he2016deep} & $\bullet$ & $\circ$ & $\circ$ & $\circ$ & $\bullet$ & $\bullet$ & $\bullet$ & $\circ$ & $\circ$ & $\circ$ & $\circ$ & $\circ$ & $\bullet$ \\
    \bottomrule
    \addlinespace[\belowrulesep]
    \multicolumn{16}{l}{\parbox{0.95\columnwidth}{\scriptsize{KEYS -- $\bullet$:~considered, $\circ$:~not considered, Sp:~spatial attention, Ch:~channel attention, Mc: masked cross-attention, Sa: self-attention, Aff.:~ affordance, C:~classification, E:~edge segmentation, S:~segmentation, L:~localisation, RN:~ResNet, RNX:~ResNeXt, RNS:~ResNeSt, SE-RNX:~squeeze and excite ResNeXt, DRN:~Dilated Residual Network, CRF:~conditioned random fields, IF:~intermediate feature maps fusion, FS:~full supervision}.}}\\
    \end{tabular}
    \label{tab:relatedworks}
\end{table}

Following instance segmentation, AffordanceNet~\cite{do2018affordancenet} and A-Mask R-CNN~\cite{caselles2021standard} modify Mask R-CNN~\cite{he2017mask} to predict the affordance masks instead of the object masks for each object localised in an image. 4C-RPN-5C~\cite{minh2020learning} and BPN~\cite{yin2022object} modify AffordanceNet to align the region of interest with the feature maps at different resolutions and predict the Intersection over Union of bounding boxes and boundaries of affordance regions. However, the additional object detection component can fail to detect objects present in the image or can cause false alarms that result in affordance regions predicted within image parts outside of actual objects. Moreover, the additional affordance edge segmentation task of BPN fails to predict precise affordance contours when edges are blurred in images or not clearly defined (e.g., occlusions or transparent objects).  To avoid the dependence from the prediction of the object location, other methods learn to independently assign each pixel of the image to an affordance class during a supervised training (per-pixel affordance segmentation)~\cite{apicella2023affordance,hussain2020fpha, gu2021visual,nguyen2016detecting,yin2022new,zhao2020object, zhang2022multi}.  
Methods such as CNN~\cite{nguyen2016detecting}, RN50-F~\cite{hussain2020fpha}, and ACANet~\cite{apicella2023affordance} can predict false positives outside the object region when objects are occluded or boundaries are not clearly defined~\cite{apicella2023affordance}.

To focus only on relevant information, recent methods belonging to both approaches use attention mechanisms to weigh feature maps extracted from the image~\cite{zhao2020object,zhang2022multi, yin2022new,yin2022object,gu2021visual}.
GSE~\cite{zhang2022multi} learns the channels weight with supervision of the affordances. 
RANet~\cite{zhao2020object} uses the additional supervision of object classes. 
DRNAtt~\cite{gu2021visual} and SEANet~\cite{yin2022object} learn similarities between channels or positions in the feature maps without direct supervision, updating the weights of the layers that compute the attention map during training. However, both DRNAtt and GSE process feature maps at low-resolutions for computational reasons, causing the feature map to degrade important details for the affordance segmentation, e.g., edges, when the object is not close to the camera. 
In RANet, occlusions can cause mistakes in the attention weights and hence a mismatch between the predicted object classes and the segmented affordances.

Unlike previous methods, classification and segmentation of affordances can be decoupled, performing the class assignment at the level of each segmentation mask~\cite{cui2023strap}. For example, STRAP is a multi-branch architecture that learns the affordance classification from the classes present in an image in one branch and the segmentation masks via weakly supervision from a point annotation of a region in another branch~\cite{tang2018regularized}. STRAP uses Conditional Random Fields to process the pixel position and color, leading to inaccurate segmentation when the object color is not clearly distinguished from the background~\cite{nguyen2016detecting, sawatzky2017weakly}. Moreover, the self-attention mechanism used in the transformer encoder processes low-resolution feature maps extracted by the backbone and may lead to lose details about the object in the image if the object scale is small. Mask2Former is a recent hybrid architecture that combines an encoder-decoder convolutional neural network with a transformer decoder to decouple the classification of classes by the segmentation, tackling different types of segmentation, e.g., semantic, instance, and panoptic segmentation~\cite{cheng2022masked}. Mask2Former introduced a masking operation in the cross-attention mechanism that combines the latent vectors with the features extracted from the image, ignoring the pixel positions outside the object region. This type of processing, not considered by previous methods, can improve the learning in tasks such affordance segmentation, in which the majority of pixels belongs to the background. In the following, we refer to the version of Mask2Former that segments affordances as M2F-AFF.

\subsection{Datasets and benchmarks}

\begin{table}[t!]
    \scriptsize
     \centering
     \caption{Characteristics of existing datasets for visual affordance segmentation.}
    \vspace{-7pt}
     \begin{tabular}{r r cc c c c c}
         \toprule
         \textbf{Dataset} & \textbf{Images} & \multicolumn{2}{c}{\textbf{Categories}} & \textbf{Real data} & \textbf{Transp.} & \textbf{TPV} & \textbf{Hand-Occ.} \\
         \cmidrule(lr){3-4} 
         & & Objects & Affordances & & & & \\
         \midrule
         AFF-Synth~\cite{christensen2022learning} & 30,245 & 21 & 7 & $\circ$ & $\circ$ & $\bullet$ & $\circ$ \\
         UMD-Synth~\cite{chu2019learning} & 37,200 & 17 & 7 & $\circ$ & $\circ$ & $\bullet$ & $\circ$ \\
         Multi-View~\cite{khalifa2022towards} & 47,210 & 37 & 15 & $\bullet$ & $\circ$ & $\bullet$ & $\circ$ \\
         HANDAL~\cite{guo2023handal} & 308,000 & 17 & 1 & $\bullet$ & $\circ$ & $\bullet$ & \raisebox{1pt}{\scalebox{0.5}{\LEFTcircle}} \\
         TRANS-AFF~\cite{jiang2022a4t} & 1,346 & 3 & 3 & $\bullet$ & $\bullet$ & $\bullet$ & $\circ$\\
         UMD~\cite{myers2015affordance} & 28,843 & 17 & 7 & $\bullet$ & $\circ$ & $\bullet$ & $\circ$ \\ 
         IIT-AFF~\cite{nguyen2017object} & 8,835 & 10 & 9 & $\bullet$ & \raisebox{1pt}{\scalebox{0.5}{\LEFTcircle}} & \raisebox{1pt}{\scalebox{0.5}{\LEFTcircle}} & \raisebox{1pt}{\scalebox{0.5}{\LEFTcircle}} \\
         CAD120-AFF~\cite{sawatzky2017weakly} & 3,090 & 11 & 6 & $\bullet$ & $\circ$ & $\bullet$ & \raisebox{1pt}{\scalebox{0.5}{\LEFTcircle}} \\ 
         FPHA-AFF~\cite{hussain2020fpha} & 4,300 & 14 & 8 & $\bullet$ & \raisebox{1pt}{\scalebox{0.5}{\LEFTcircle}} & $\circ$ & $\bullet$ \\
         CHOC-AFF~\cite{apicella2023affordance} & 138,240 & 3 & 3 & \raisebox{1pt}{\scalebox{0.5}{\LEFTcircle}} & \raisebox{1pt}{\scalebox{0.5}{\LEFTcircle}} & $\bullet$ & \raisebox{1pt}{\scalebox{0.5}{\LEFTcircle}} \\
         \bottomrule \addlinespace[\belowrulesep]
         \multicolumn{8}{l}{\parbox{0.7\columnwidth}{\scriptsize{KEYS -- Transp.:~transparency, TPV:~third person view, Hand-Occ.:~hand-occlusion, \\ $\bullet$: considered, $\circ$: not considered, \raisebox{1pt}{\scalebox{0.4}{\LEFTcircle}}: mixed (partially considered)}}}
     \end{tabular}
     \label{tab:affdatasets}
 \end{table}

Table~\ref{tab:affdatasets} shows the main characteristics of existing datasets for affordance segmentation. Methods are usually compared and benchmarked on the testing split of datasets with images of real scenarios. In general, existing benchmarks do not have a complete overlap of affordance classes, hence benchmarking the performance of methods trained using different datasets is impossible.  

Some of the training datasets are synthetic i.e., created using computer graphics, due to the need of a large number of images~\cite{apicella2023affordance, chu2019learning, christensen2022learning}. These datasets are annotated by labelling the regions of CAD models, rendering the CAD model in a virtual scene, and then projecting back the new annotation in the camera frame. This type of segmentation annotation has a perfect spatial overlapping with the object rendered by the virtual camera. UMD-Synth~\cite{chu2019learning}  has objects in different poses placed on a tabletop while varying colors and texture. AFF-Synth~\cite{christensen2022learning} has images generated using domain randomization techniques to overcome the gap between simulated and real data. Unlike UMD-Synth, each image of AFF-Synth includes multiple objects and some of them are used as distractors during training as they do not have affordance annotation. The main issue with synthetic datasets is the sim-to-real gap in performance: models trained on completely synthetic datasets tend to have poor generalisation performance to real data. 

In real datasets images are captured by a real camera and manually annotated to label the pixels of the object regions with an affordance class, following the convention on classes names and definition established by the dataset creators~\cite{christensen2022learning, chu2019learning, jiang2022a4t, khalifa2022towards, myers2015affordance, nguyen2017object, sawatzky2017weakly}. In general, annotators label object regions with the action an agent can perform \textit{with} or \textit{on} that region (e.g., the handle of an hammer is \textit{graspable} while the hard top allows to \textit{pound} other objects). Manual annotation is a time consuming task, requires a certain amount of effort, and is prone to errors and imperfections. For example, some objects in the scene might not be annotated, especially in the case of clutter or small object scale. Moreover, the mask annotation might be incomplete or outside the object boundaries, depending on the precision of annotators in distinguishing between regions of the object and the background. 
During training, these mistakes can influence the models learning, and during evaluation, the ranking and values of the performance measure. A model that has imperfect outputs, but more similar to the annotation error, will obtain a higher score than a model predicting masks that are visually better than the annotated mask. 


UMD~\cite{myers2015affordance} is composed by images of objects placed on a blue rotating table in the same environment. Similarly, Multi-View~\cite{khalifa2022towards} contains images of objects placed on a white rotating table keeping the same lighting. Multi-view has a higher number of objects and affordances than UMD, however the training set consists only of object crops, hence models cannot be trained using an entire scene. HANDAL~\cite{guo2023handal} varies objects placement in outdoor and indoor environments, such as a street or a living room, which also contain clutter. To annotate images, HANDAL uses a hybrid approach between manual and automatic annotation based on recent off-the-shelf methods. BundleSDF predicts the 6D pose of the objects in each frame and reconstruct the CAD model~\cite{wen2023bundlesdf}, then the annotator labels the handle of the CAD models as \textit{graspable}. The final annotation mask is obtained by projecting the annotated CAD model in the camera frame through the estimated object and camera pose. The main limitation of HANDAL is the focus on objects having a handle, limiting the application of trained models to other types of object. TRANS-AFF~\cite{jiang2022a4t} is focused on few categories of transparent objects placed on a tabletop, increasing the difficulty to predict affordances, as objects are also in different poses. IIT-AFF~\cite{nguyen2017object} is composed by images of objects placed in a cluttered scene to better reflect a scenario with clutter and occlusions. One part of the dataset is collected from other datasets (e.g., ImageNet~\cite{deng2009imagenet}) by varying the instances and the setting of the images. In CAD120-AFF~\cite{sawatzky2017weakly}, images are sampled from videos of humans performing activities in a realistic setting, e.g., kitchen or office, with more than one object in the scene and also with hand occlusions. FPHA-AFF~\cite{hussain2020fpha} has images of hand-held objects acquired from an egocentric point of view. However, the affordance annotation of this dataset is currently not publicly available and egocentric images contains arms from the bottom of the image, resulting in objects highly occluded by the hands. CHOC-AFF~\cite{apicella2023affordance} is the only available dataset with a high number of images (more than 129,000) with hand-occluded objects, varying hand and object pose. 

Table~\ref{tab:umdtrainingsetup_pre} highlights the main limitation of previous works when training models on UMD. The different setups used by the methods make their direct comparisons unfair~\cite{do2018affordancenet, gu2021visual, zhang2022multi}.
In this work, we tackle affordance segmentation of a single object in two scenarios, on a tabletop and hand-occluded, and revisit the UMD and CHOC-AFF benchmarks. We select these two datasets due to the trade-off between real images of the entire scene, number of images, scenario, number of object and affordance categories.

\begin{table}[t!]
    \centering
    \scriptsize
    \setlength\tabcolsep{3pt}
    \caption{Training setup of affordance segmentation methods for the UMD dataset~\cite{myers2015affordance}. Our setup is applied to all models to ensure a fair comparison.}
    \vspace{-7pt}
    \begin{tabular}{c c c c c c c c}
    \toprule
    \textbf{Training setup} & \textbf{Image resolution} & \multicolumn{4}{c}{\textbf{Data augmentation}} & \multicolumn{2}{c}{\textbf{Image resize procedure}} \\
    \cmidrule(lr){3-6}\cmidrule(lr){7-8}
    & & Flipping & Scaling & Rotating & Color jittering & Training set & Testing set\\
    \midrule
    ~\cite{nguyen2016detecting} & $320 \times 240$ & $\circ$ & $\circ$ & $\circ$ & $\circ$ & centre-crop & sliding window \\    
    ~\cite{do2018affordancenet} & $1000 \times 600$ & $\circ$ & $\circ$ & $\circ$ & $\circ$ & unknown & unknown \\ 
    ~\cite{zhang2022multi} & $400 \times 400$ & $\bullet$ & $\bullet$ & $\circ$ & $\circ$ & crop & unknown \\ 
    ~\cite{zhao2020object} & $224 \times 224$ & $\circ$ & $\circ$ & $\circ$ & $\circ$ & centre-crop & unknown \\
    ~\cite{yin2022object} & $1000 \times 600$  & $\bullet$ & $\bullet$ & $\bullet$ & $\bullet$ & unknown & unknown \\ 
    ~\cite{gu2021visual} & $320 \times 240$ & $\circ$ & $\circ$ & $\circ$ & $\circ$ & centre-crop & unknown \\ 
    \rowcolor{lightgray}
    Ours & $640 \times 480$ & $\bullet$ & $\bullet$ & $\circ$ & $\bullet$ & centre-crop & no cropping \\
    \bottomrule
    \addlinespace[\belowrulesep]
    \multicolumn{7}{l}{\parbox{0.7\linewidth}{\scriptsize{KEYS -- $\bullet$:~considered, $\circ$:~not considered}}}\\
    \end{tabular}
    \label{tab:umdtrainingsetup_pre}
\end{table}

\section{Unoccluded object on a tabletop}

\subsection{Experimental setup} 

We consider UMD~\cite{myers2015affordance}, a standard dataset used to evaluate the visual affordance segmentation of a single object without occlusions. UMD has 28,843 images of tools and containers, each placed on a blue rotating table, and acquired from a fixed view with the object in the center~\cite{myers2015affordance}. 
Object instances are not evenly distributed among the 17 object categories, e.g., there are 10 instances of spoon and 2 instances of pot. UMD is annotated with 7 affordance classes: \textit{grasp}, \textit{cut}, \textit{scoop}, \textit{contain}, \textit{pound}, \textit{support}, \textit{wrap-grasp}. UMD provides a pre-defined split of the dataset into training ($14,823$ images) and testing sets ($14,020$ images), holding out approximately half of object instances per category. The training set is subsequently split into training and validation sets, sampling $30\%$ of the images for the latter.

We compare M2F-AFF with three existing methods: AffordanceNet~\cite{do2018affordancenet}, CNN~\cite{nguyen2016detecting} and DRNAtt~\cite{gu2021visual}. CNN is based on an encoder-decoder architecture to segment affordances. AffordanceNet is a two-stage method that detects the object and segments affordances. DRNAtt is an attention-based architecture and is the best-performing method in tabletop settings~\cite{gu2021visual}. Due to the unavailable implementations, we re-implemented DRNAtt and CNN, and trained all the methods using the same experimental setup to ensure fair conditions. 

Previous works evaluated methods performance using $F^w_\beta$~\cite{margolin2014evaluate} as performance measure. $F^w_\beta$ associates a different weight to the prediction errors based on the Euclidean distance to the annotated mask. Despite the positive aspect of attributing different importance to prediction mistakes, $F^w_\beta$ was proposed to evaluate foreground maps and therefore ignores the classes that are not present in the annotation. To also count prediction mistakes of classes that are not in the annotation, and when comparing models predictions $\{\hat{M}_i\}^{N}_{i=1}$ with annotations $\{M_i\}^{N}_{i=1}$, where $N$ is the number of images in a dataset, we use per-class Jaccard index ($J$) to measure how many pixels predicted as class $s$ are correct, among all pixels (predicted and annotated):
\begin{equation}\label{eq:jaccard}
    J = \frac{\sum^{N}_{i=1} \sum^{W \times H}_{x} TP}{\sum^{N}_{i=1} \sum^{W \times H}_{x} TP + FP + FN} \: ,
\end{equation}
where we consider a pixel $x$ having value $s$ in the prediction $\hat{M}$ and in the annotation $M$ as true positive ($TP$), a pixel having value $s$ in $\hat{M}$, but not in $M$ as false positive ($FP$), a pixel having value $s$ in $M$, but not in $\hat{M}$ as false negative ($FN$).

\subsection{Strategy and parameters setting}\label{ssec:parameters}

Table~\ref{tab:umdtrainingsetup_pre} compares the training setups of the previous works with ours. Although almost each paper used a different setup for training the method, some previous works made straight comparisons of performance values with the results in related works~\cite{do2018affordancenet, gu2021visual, zhang2022multi}. CNN~\cite{nguyen2016detecting}, GSE~\cite{zhang2022multi}, and RANet~\cite{zhao2020object} use an input resolution lower than the UMD images ($640 \times 480$), cropping the images around the annotation to fit the resolution. This procedure implies the presence of a perfect object detector to locate the object in the scene. Unlike these setups, ours uses the full image resolution for every model, avoiding the perfect detector assumption for all the models. Methods that use the additional object detection task, such as AffordanceNet~\cite{do2018affordancenet} and BPN~\cite{yin2022object}, resize the image to a resolution with minimum height $600$ pixels and maximum with $1,000$ pixels. Thus, AffordanceNet and BPN process feature maps with higher resolutions than the ones obtained with $640 \times 480$ images (UMD resolution). Only GSE~\cite{zhang2022multi} and BPN~\cite{yin2022object} use augmentation to limit the overfitting on the images having all similar background, i.e., the blue rotating table. In this work, we train CNN and DRNAtt using cross-entropy loss, Adam optimiser and batch size $4$. For AffNet, we used a combination of cross-entropy and smooth-L1 losses for detection and cross-entropy for segmentation, mini-batch gradient descent with weight decay $0.001$ and batch size $2$ to fit our available GPU memory. We initialise all the mentioned architectures with a learning rate of $10^{-3}$. We train M2F-AFF with AdamW optimiser~\cite{loshchilov2017decoupled} with batch size $4$ and learning rate $10^{-4}$. We follow Mask2Former setup~\cite{cheng2022masked}, using hungarian algorithm~\cite{kuhn1955hungarian} to match the prediction from each latent vector with the corresponding annotation. We minimize the linear combination among the cross-entropy for classification $L_{cls}$, with the binary dice loss $L_{dice}$ and the binary cross-entropy $L_{ce}$ for segmentation. Losses are weighted using hyperparameters  $\lambda_{ce}$, $\lambda_{dice}$, and $\lambda_{cls}$.

For all models, we decrease the learning rate by a factor of $0.5$, if there is no increase of the mean Intersection over Union in the validation set for $3$ consecutive epochs. We set the maximum number of epochs to $100$. We also use early stopping with a patience of $5$ epochs to reduce over-fitting. We use flipping with a probability of $0.5$, scaling by randomly sampling the scale factor in the interval $[1, 1.5]$ and center-cropping to simulate a zoom-in effect, color jitter with brightness, contrast, and saturation sampled randomly in the interval $[0.9, 1.1]$, and hue sampled randomly in the interval $[-0.1, 0.1]$. These augmentations allow to increase variability in the training set. We initialise all backbones with weights trained on ImageNet~\cite{deng2009imagenet}.

\subsection{Implementation details}\label{ssec:implementation_details}
For DRNAtt and CNN, we used the backbones described in the corresponding paper due to design requirements. In DRNAtt, DRN22 backbone allows to maintain higher resolution compared to a ResNet backbone. In CNN, the features down-sampling is not performed by VGG16 blocks but by down-sampling layers and the indices of down-sampled position are saved to be re-used in the decoder, while a ResNet backbone decreases the resolution using stride $2$ in convolutional filters. We used an available implementation of AffordanceNet~\cite{christensen2022learning}, that differs from the original one by implementing the ResNet-50 backbone with the Feature Pyramid Network instead of using a VGG backbone. Moreover, to fit the available memory, each up-sampling layer in the segmentation decoder contains $128$ channels as opposed to $512$ in AffordanceNet. The first up-sampling layers has kernel size $4$ instead of $8$ and the stride is $2$ instead of $4$. For M2F-AFF, we chose the Mask2Former version that uses the ResNet-50 backbone, multi-scale deformable attention~\cite{zhu2020deformable}, 3 decoder layers repeating the sequence of a masked cross-attention layer with 8 heads, a self-attention layer with 8 heads, a feed-forward layer, and 100 queries. During training, we follow Mask2Former setup, setting the default loss weights $\lambda_{ce}=5$, $\lambda_{dice}=5$, $\lambda_{cls}=2$ and to compute losses values, we sample sets of $K=12,544$ points for prediction and ground truth masks~\cite{cheng2022masked}. During inference, we use a confidence threshold of $0.5$ to separate background from an affordance region.

\subsection{Comparative analysis and segmentation results} 
Fig.~\ref{fig:res_all_umd} compares the performance of affordance segmentation methods on the UMD testing set using the Jaccard Index. On average, M2F-AFF outperforms other methods. Previous works simply reported the $F^w_\beta$ even though the experimental setup used was different. The ranking of methods is similar to the one in previous works~\cite{gu2021visual, chen2023survey}, but with a different range of performance values ($F^w_\beta$ results in the Supplementary Material document). DRNAtt is the previous work that has the best performance compared to CNN and AffordanceNet, especially for the class \textit{support}. M2F-AFF has the highest improvement in Jaccard Index for the class \textit{cut} ($+13$ percentage points) and \textit{scoop} ($+16$ percentage points) compared to DRNAtt.

\pgfplotstableread{data/fwbeta_re_implemented_umd.txt}\fwbeta
\pgfplotstableread{data/jaccard_re_implemented_umd.txt}\jaccard
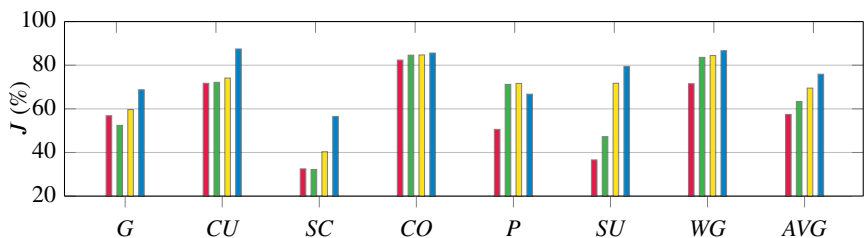
\begin{figure}[t!]
    \centering
    \begin{tikzpicture}
    \begin{axis}[
    axis x line*=bottom,
    axis y line*=left,
    enlarge x limits=false,
    ybar,
    width=\linewidth,
    bar width=2.1pt,
    typeset ticklabels with strut, 
    xmin=-1,xmax=38.5,
    xtick={2, 6.8, 11.6, 16.4, 21.2, 26, 30.8, 35.6},
    xticklabels={$\textit{G}$, $\textit{CU}$, $\textit{SC}$, $\textit{CO}$, $\textit{P}$, $\textit{SU}$, $\textit{WG}$, $\textit{AVG}$},
    height=0.32\linewidth,
    ymin=20,ymax=100,
    ytick={20,40,60,80,100},
    ylabel={$J$ (\%)},
    ylabel style={yshift=-10pt},
    label style={font=\footnotesize},
    tick label style={font=\footnotesize},
    ymajorgrids=true,
    ]
    \addplot+[ybar, black, fill=dc1, draw opacity=0.5] table[x=CoID,y=CNN]{\jaccard};
    \addplot+[ybar, black, fill=dc2, draw opacity=0.5] table[x=CoID,y=AffordanceNet]{\jaccard};
    \addplot+[ybar, black, fill=dc3, draw opacity=0.5] table[x=CoID,y=DRNAtt]{\jaccard};
    \addplot+[ybar, black, fill=dc4, draw opacity=0.5] table[x=CoID,y=MtwoF]{\jaccard};
    \end{axis}
    \begin{axis}[ 
        axis x line*=top,
        axis y line*=right,
        width=\linewidth,
        height=0.32\linewidth,
        tick label style={font=\footnotesize, align=center,text width=3cm}, 
        xmin=-0.5,xmax=38,
        xtick={2, 6.8, 11.6, 16.4, 21.2, 26, 30.8, 35.6, 35.6},
        xticklabels={},
        typeset ticklabels with strut,
        label style={font=\footnotesize},
        ymin=20,ymax=100,
        yticklabels={},
    ]
    \end{axis}
    \end{tikzpicture}
    \caption{Comparison of segmentation accuracy between methods on the testing set of UMD~\cite{myers2015affordance}. Methods are re-implemented and trained using the same experimental setup.
   KEYS -- G: \textit{grasp}, CU: \textit{cut}, SC: \textit{scoop}, CO: \textit{contain}, P: \textit{pound}, SU: \textit{support}, WG: \textit{wrap-grasp}, AVG: \textit{average},
    \protect\raisebox{2pt}{\protect\tikz \protect\draw[dc1,line width=2] (0,0) -- (0.3,0);}~\textit{CNN}~\cite{nguyen2016detecting},
    \protect\raisebox{2pt}{\protect\tikz \protect\draw[dc2,line width=2] (0,0) -- (0.3,0);}~\textit{AffordanceNet}~\cite{do2018affordancenet},
    \protect\raisebox{2pt}{\protect\tikz \protect\draw[dc3,line width=2] (0,0) -- (0.3,0);}~\textit{DRNAtt}~\cite{gu2021visual}, 
    \protect\raisebox{2pt}{\protect\tikz \protect\draw[dc4,line width=2] (0,0) -- (0.3,0);}~\textit{M2F-AFF}.
   }
    \label{fig:res_all_umd}
\end{figure}

\begin{figure}[t!]
    \centering
    \includegraphics[width=\linewidth]{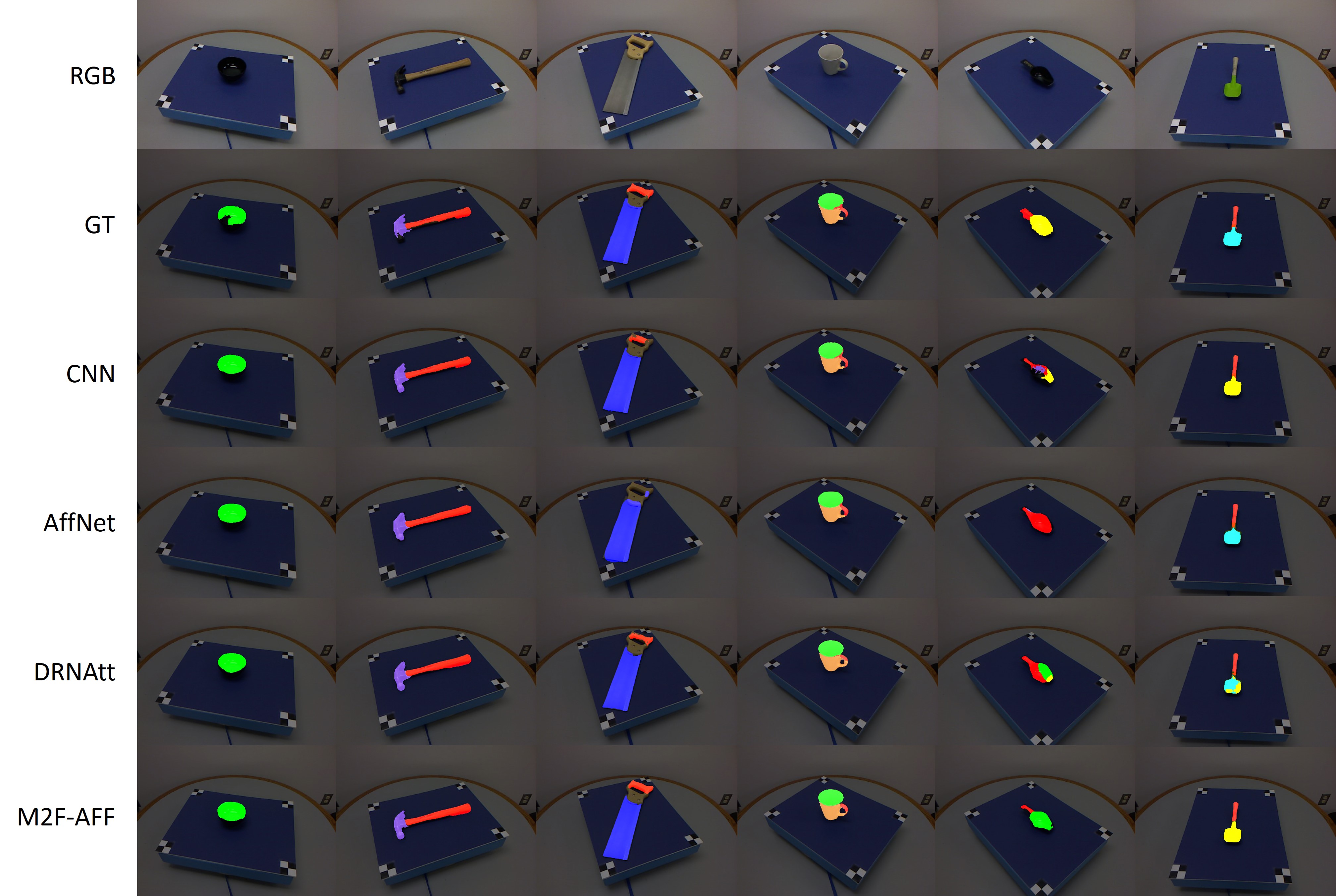}
    \caption{Comparison between the ground truth and models prediction on the UMD testing set. 
    Legend: \protect\raisebox{2pt}{\protect\tikz \protect\draw[tgraspable,line width=2] (0,0) -- (0.3,0);}~\textit{graspable},
    \protect\raisebox{2pt}{\protect\tikz \protect\draw[tcut,line width=2] (0,0) -- (0.3,0);}~\textit{cut},
    \protect\raisebox{2pt}{\protect\tikz \protect\draw[tscoop,line width=2] (0,0) -- (0.3,0);}~\textit{scoop},
    \protect\raisebox{2pt}{\protect\tikz \protect\draw[tcontain,line width=2] (0,0) -- (0.3,0);}~\textit{contain},
    \protect\raisebox{2pt}{\protect\tikz \protect\draw[tpound,line width=2] (0,0) -- (0.3,0);}~\textit{pound},
    \protect\raisebox{2pt}{\protect\tikz \protect\draw[tsupport,line width=2] (0,0) -- (0.3,0);}~\textit{support},
    \protect\raisebox{2pt}{\protect\tikz \protect\draw[twgrasp,line width=2] (0,0) -- (0.3,0);}~\textit{wrap-grasp}}
    \label{fig:qual_res_umd}
\end{figure}

Fig.~\ref{fig:qual_res_umd} compares the predictions of methods with the affordance segmentation annotations. For some images, the annotation of the affordance mask is  either wider or smaller than the actual region in the image (see first two columns). In these cases, models visually segments the affordance regions more accurately than the annotation. For example, the predicted \textit{pound} region covers the whole top of the hammer (second column). For other cases, such as the images of a saw and a mug, both annotation and M2F-AFF prediction are similar (see third and fourth columns). Furthermore, M2F-AFF predictions are visually accurate despite class confusions (last two columns), even compared to methods that segment the correct class (AffNet in the last column).

\section{Hand-held containers}

\subsection{Dataset}

We follow the experimental setup in previous evaluations~\cite{apicella2023affordance} and we report the results from their validation. This setup includes a mixed-reality dataset and two testing sets of real images, all publicly available; Jaccard index as performance measures (see Eq.~\ref{eq:jaccard}); and four methods, all trained on the mixed-reality dataset and evaluated on both the mixed-reality dataset and the two testing sets. 


CORSMAL Hand-Occluded Containers Affordance (CHOC-AFF)~\cite{apicella2023affordance,weber2022mixed} has $138,240$ images of 48 synthetic containers hand-held by synthetic hands rendered on top of 30 real backgrounds, and with annotations of the object affordances (\textit{graspable} and \textit{contain}) and forearm masks (\textit{arm}). 
The locations and poses of the hand-held objects vary across images, with 129,600 images of hand-held objects rendered above a table, and 8,640 images with objects rendered on a tabletop. CHOC-AFF has 
four splits: a training set of 103,680 images, combining 26 backgrounds and 36 objects instances; a validation set of 17,280 images, using all 30 backgrounds and holding out 6 objects instances; a testing set of 13,824 images (CHOC-B), used to assess the generalisation performance to backgrounds never seen during training; and another testing set of 17,280 images (CHOC-I), to assess the generalisation performance to object instances never seen during training.


HO-3D-AFF and CCM-AFF are the two testing sets of real images used to assess the generalisation performance of the models~\cite{apicella2023affordance}. HO-3D-AFF sampled 150 images from HO-3D~\cite{hampali2020honnotate}, a dataset with images of hands holding and manipulating objects, such as tools and containers (only mugs, boxes, and cans, in different poses, were selected for the sampled images). HO-3D-AFF complemented the existing annotations with affordance masks. CCM-AFF sampled 150 images from CORSMAL Containers Manipulation (CCM)~\cite{xompero2020corsmal}, a dataset consisting of audio-visual recordings of people manipulating containers. Images were selected from a lateral view, while varying object instances and conditions, such as  lighting, filling type and amount, and presence of the tablecloth. CCM-AFF annotated the affordance and forearm masks. Both testing sets use the same classes of CHOC-AFF:  \textit{graspable}, \textit{contain}, and \textit{arm}. 

\subsection{Methods under comparison} 

We compare the performance of our re-implemented versions of RN50-F~\cite{hussain2020fpha}, DRNAtt~\cite{gu2021visual}, RN18-U~\cite{ronneberger2015u}, ACANet~\cite{apicella2023affordance} with M2F-AFF trained with CHOC-AFF affordance classes. RN50-F uses a ResNet-50 encoder with a pyramid scene parsing module~\cite{zhao2017pyramid} to segment only the object affordances \textit{graspable} and \textit{contain}~\cite{hussain2020fpha}. DRNAtt uses position and channel attention mechanisms in parallel after the feature extraction stage~\cite{gu2021visual}. The outputs of the attention modules are summed element-wise and the result is up-sampled through a learnable decoder. RN18-U and ACANet~\cite{apicella2023affordance} are UNet-like models that gradually down-sample feature maps in the encoder and up-sample them in the decoder, preserving the information via skip connections~\cite{ronneberger2015u}. Both models implement a ResNet encoder~\cite{he2016deep}. ACANet separately segments object and hand regions, using these masks to weigh the feature maps learnt in a third branch for the final affordance segmentation. Additionally, we trained a version of ACANet with ResNet-50 to have the same backbone of M2F-AFF and RN50-F. 

\pgfplotstableread{data/jaccard_handoccluded_CHOCB.txt}\jaccardchocb
\pgfplotstableread{data/jaccard_handoccluded_CHOCI.txt}\jaccardchoci
\pgfplotstableread{data/jaccard_handoccluded_HO3D.txt}\jaccardhod
\pgfplotstableread{data/jaccard_handoccluded_CCM.txt}\jaccardccm
\begin{figure}[t!]
    \centering
    \begin{subfigure}[b]{0.5\linewidth}
    \centering
    \begin{tikzpicture}
    \begin{axis}[
    axis x line*=bottom,
    axis y line*=left,
    enlarge x limits=false,
    ybar,
    width=1.1\linewidth,
    bar width=2.1pt,
    typeset ticklabels with strut, 
    xmin=-0,xmax=8,
    xtick={1, 3, 5, 7},
    xticklabels={$\textit{G}$, $\textit{CO}$, $\textit{A}$, $\textit{AVG}$},
    height=0.58\linewidth,
    ymin=60,ymax=100,
    ytick={60,70,80,90,100},
    ylabel={$J$ (\%)},
    ylabel style={yshift=-10pt},
    label style={font=\footnotesize},
    tick label style={font=\scriptsize},
    ymajorgrids=true,
    title={\textbf{CHOC-B}},
    title style={font=\footnotesize, yshift=-6pt}
    ]
    \addplot+[ybar, black, fill=dc7, draw opacity=0.5] table[x=CoID,y=RN50F]{\jaccardchocb};
    \addplot+[ybar, black, fill=dc3, draw opacity=0.5] table[x=CoID,y=DRNAtt]{\jaccardchocb};
    \addplot+[ybar, black, fill=dc10, draw opacity=0.5] table[x=CoID,y=RN18U]{\jaccardchocb};
    \addplot+[ybar, black, fill=dc13, draw opacity=0.5] table[x=CoID,y=ACANet]{\jaccardchocb};
    \addplot+[ybar, black, fill=dc17, draw opacity=0.5] table[x=CoID,y=ACANet50]{\jaccardchocb};
    \addplot+[ybar, black, fill=dc4, draw opacity=0.5] table[x=CoID,y=M2Former]{\jaccardchocb};
    \end{axis}
    \begin{axis}[ 
        axis x line*=top,
        axis y line*=right,
        width=1.1\linewidth,
        height=0.58\linewidth,
        tick label style={font=\footnotesize, align=center,text width=3cm}, 
        xmin=-0,xmax=8,
        xtick={2, 6.8, 11.6, 16.4},
        xticklabels={},
        typeset ticklabels with strut,
        label style={font=\footnotesize},
        ymin=60,ymax=100,
        ytick={60,70,80,90,100},
        yticklabels={}
    ]
    \end{axis}
    \end{tikzpicture}
    \end{subfigure}
    \hspace{-30pt}
    \begin{subfigure}[b]{0.5\linewidth}
    \centering
    \begin{tikzpicture}
    \begin{axis}[
    axis x line*=bottom,
    axis y line*=left,
    enlarge x limits=false,
    ybar,
    width=1.1\linewidth,
    bar width=2.1pt,
    typeset ticklabels with strut, 
    xmin=-0,xmax=8,
    xtick={1, 3, 5, 7},
    xticklabels={$\textit{G}$, $\textit{CO}$, $\textit{A}$, $\textit{AVG}$},
    height=0.58\linewidth,
    ymin=60,ymax=100,
    ytick={60,70,80,90,100},
    yticklabels={},
    label style={font=\footnotesize},
    tick label style={font=\scriptsize},
    ymajorgrids=true,
    title={\textbf{CHOC-I}},
    title style={font=\footnotesize, yshift=-6pt}
    ]
    \addplot+[ybar, black, fill=dc7, draw opacity=0.5] table[x=CoID,y=RN50F]{\jaccardchoci};
    \addplot+[ybar, black, fill=dc3, draw opacity=0.5] table[x=CoID,y=DRNAtt]{\jaccardchoci};
    \addplot+[ybar, black, fill=dc10, draw opacity=0.5] table[x=CoID,y=RN18U]{\jaccardchoci};
    \addplot+[ybar, black, fill=dc13, draw opacity=0.5] table[x=CoID,y=ACANet]{\jaccardchoci};
    \addplot+[ybar, black, fill=dc17, draw opacity=0.5] table[x=CoID,y=ACANet50]{\jaccardchoci};
    \addplot+[ybar, black, fill=dc4, draw opacity=0.5] table[x=CoID,y=M2Former]{\jaccardchoci};
    \end{axis}
    \begin{axis}[ 
        axis x line*=top,
        axis y line*=right,
        width=1.1\linewidth,
        height=0.58\linewidth,
        tick label style={font=\footnotesize, align=center,text width=3cm}, 
        xmin=-0,xmax=8,
        xtick={1, 3, 5, 7},
        xticklabels={},
        typeset ticklabels with strut,
        label style={font=\footnotesize},
        ymin=60,ymax=100,
        ytick={60,70,80,90,100},
        yticklabels={},
    ]
    \end{axis}
    \end{tikzpicture}
    \end{subfigure}
    \begin{subfigure}[b]{0.5\linewidth}
    \centering
    \begin{tikzpicture}
    \hspace{-3pt}
    \begin{axis}[
    axis x line*=bottom,
    axis y line*=left,
    enlarge x limits=false,
    ybar,
    width=1.1\linewidth,
    bar width=2.1pt,
    typeset ticklabels with strut, 
    xmin=-0,xmax=8,
    xtick={1, 3, 5, 7},
    xticklabels={$\textit{G}$, $\textit{CO}$, $\textit{A}$, $\textit{AVG}$},
    height=0.58\linewidth,
    ymin=0,ymax=80,
    ytick={0, 20, 40, 60, 80},
    ylabel={$J$ (\%)},
    ylabel style={yshift=-5pt},
    label style={font=\footnotesize},
    tick label style={font=\scriptsize},
    ymajorgrids=true,
    title={\textbf{HO-3D-AFF}},
    title style={font=\footnotesize, yshift=-6pt}
    ]
    \addplot+[ybar, black, fill=dc7, draw opacity=0.5] table[x=CoID,y=RN50F]{\jaccardhod};
    \addplot+[ybar, black, fill=dc3, draw opacity=0.5] table[x=CoID,y=DRNAtt]{\jaccardhod};
    \addplot+[ybar, black, fill=dc10, draw opacity=0.5] table[x=CoID,y=RN18U]{\jaccardhod};
    \addplot+[ybar, black, fill=dc13, draw opacity=0.5] table[x=CoID,y=ACANet]{\jaccardhod};
    \addplot+[ybar, black, fill=dc17, draw opacity=0.5] table[x=CoID,y=ACANet50]{\jaccardhod};
    \addplot+[ybar, black, fill=dc4, draw opacity=0.5] table[x=CoID,y=M2Former]{\jaccardhod};
    \end{axis}
    \begin{axis}[ 
        axis x line*=top,
        axis y line*=right,
        width=1.1\linewidth,
        height=0.58\linewidth,
        tick label style={font=\footnotesize, align=center,text width=3cm}, 
        xmin=-0,xmax=8,
        xtick={1, 3, 5, 7},
        xticklabels={},
        typeset ticklabels with strut,
        label style={font=\footnotesize},
        ymin=0,ymax=80,
        ytick={0, 20, 40, 60, 80},
        yticklabels={},
    ]
    \end{axis}
    \end{tikzpicture}
    \end{subfigure}
    \hspace{-30pt}
    \begin{subfigure}[b]{0.5\linewidth}
    \centering
    \begin{tikzpicture}
    \begin{axis}[
    axis x line*=bottom,
    axis y line*=left,
    enlarge x limits=false,
    ybar,
    width=1.1\linewidth,
    bar width=2.1pt,
    typeset ticklabels with strut, 
    xmin=-0,xmax=8,
    xtick={1, 3, 5, 7},
    xticklabels={$\textit{G}$, $\textit{CO}$, $\textit{A}$, $\textit{AVG}$},
    height=0.58\linewidth,
    ymin=0,ymax=80,
    ytick={0, 20, 40, 60, 80},
    yticklabels={},
    label style={font=\footnotesize},
    tick label style={font=\scriptsize},
    ymajorgrids=true,
    title={\textbf{CCM-AFF}},
    title style={font=\footnotesize, yshift=-6pt}
    ]
    \addplot+[ybar, black, fill=dc7, draw opacity=0.5] table[x=CoID,y=RN50F]{\jaccardccm};
    \addplot+[ybar, black, fill=dc3, draw opacity=0.5] table[x=CoID,y=DRNAtt]{\jaccardccm};
    \addplot+[ybar, black, fill=dc10, draw opacity=0.5] table[x=CoID,y=RN18U]{\jaccardccm};
    \addplot+[ybar, black, fill=dc13, draw opacity=0.5] table[x=CoID,y=ACANet]{\jaccardccm};
    \addplot+[ybar, black, fill=dc17, draw opacity=0.5] table[x=CoID,y=ACANet50]{\jaccardccm};
    \addplot+[ybar, black, fill=dc4, draw opacity=0.5] table[x=CoID,y=M2Former]{\jaccardccm};
    \end{axis}
    \begin{axis}[ 
        axis x line*=top,
        axis y line*=right,
        width=1.1\linewidth,
        height=0.58\linewidth,
        tick label style={font=\footnotesize, align=center,text width=3cm}, 
        xmin=-0,xmax=8,
        xtick={1, 3, 5, 7},
        xticklabels={},
        typeset ticklabels with strut,
        label style={font=\footnotesize},
        ymin=0,ymax=80,
        ytick={0, 20, 40, 60, 80},
        yticklabels={},
    ]
    \end{axis}
    \end{tikzpicture}
    \end{subfigure}
    \caption{Comparison of the affordance and arm segmentation results between the models on the two mixed-reality testing sets (top row) and on the two real testing sets (bottom row). Note the differnt y-axis limits. KEYS -- G: \textit{grasp}, CO: \textit{contain}, A: \textit{arm}, AVG: \textit{average},
    \protect\raisebox{2pt}{\protect\tikz \protect\draw[dc7,line width=2] (0,0) -- (0.3,0);}~\textit{RN50F}~\cite{hussain2020fpha},
    \protect\raisebox{2pt}{\protect\tikz \protect\draw[dc3,line width=2] (0,0) -- (0.3,0);}~\textit{DRNAtt}~\cite{gu2021visual},
    \protect\raisebox{2pt}{\protect\tikz \protect\draw[dc10,line width=2] (0,0) -- (0.3,0);}~\textit{RN18-U}~\cite{apicella2023affordance}, 
    \protect\raisebox{2pt}{\protect\tikz \protect\draw[dc13,line width=2] (0,0) -- (0.3,0);}~\textit{ACANet}~\cite{apicella2023affordance},
    \protect\raisebox{2pt}{\protect\tikz \protect\draw[dc17,line width=2] (0,0) -- (0.3,0);}~\textit{ACANet50}~\cite{apicella2023affordance}, 
    \protect\raisebox{2pt}{\protect\tikz \protect\draw[dc4,line width=2] (0,0) --(0.3,0);}~\textit{M2F-AFF}~\cite{cheng2022masked}. 
   }
    \label{fig:handoccluded_resall}
\end{figure}
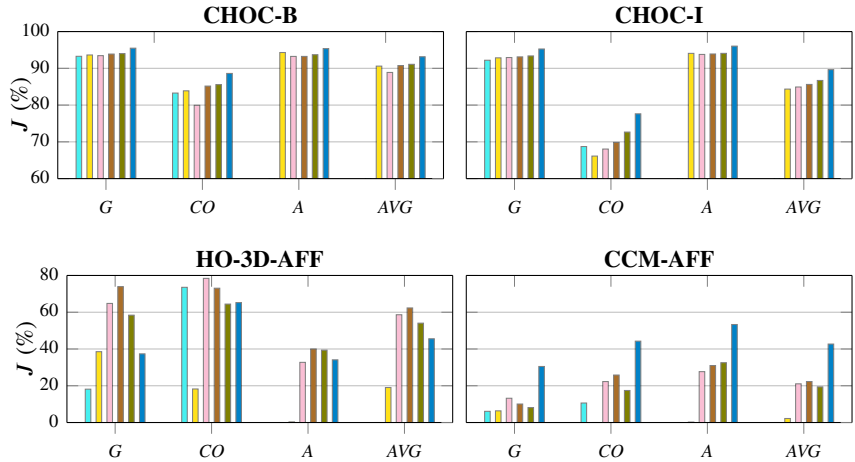

M2F-AFF uses the same architecture described in Sec.~\ref{ssec:parameters} and ~\ref{ssec:implementation_details}, only the number of output classes changes to fit the hand-occluded settings. We used the same training setup of compared methods~\cite{apicella2023affordance}, initialising the backbone with ImageNet weights. We use early stopping with a patience of $10$ epochs, setting the maximum number of epochs to $100$. The learning rate decreases by a factor of $0.5$ if there is no increase of the mean Intersection over Union in the validation set for $3$ consecutive epochs. We use the cropping window technique described in the previous work~\cite{apicella2023affordance}, resize images by a factor sampled in the interval $[1,1.5]$ and center crop the resized image with a $480 \times 480$ window. To simulate the other arm holding the object, we use horizontal flip with a probability of $0.5$. M2F-AFF uses batch size 4, learning rate $0.0001$, and an additional Gaussian noise augmentation with variance in range $[10, 100]$ to increase variability in training data.

\subsection{Segmentation results under hand-occlusions} 

Fig.~\ref{fig:handoccluded_resall} compares the results of M2F-AFF and previous methods. Models achieve high and similar performance on the testing sets of CHOC-AFF, whereas there are significant performance differences across the models on the two testing sets with real images. In HO-3D-AFF, the best-performing model is still ACANet, whereas M2F-AFF performance is mostly affected by the missing segmentation of \textit{graspable} regions of the boxes close to the camera. 
This result suggests that the model segments better affordances of objects with low pixel occupancy. CCM-AFF is more challenging than HO-3D-AFF due to the presence of the human body and transparent containers. M2F-AFF outperforms existing models on this testing set.  

\begin{figure}[t!]
    \centering
    \includegraphics[width=\linewidth]{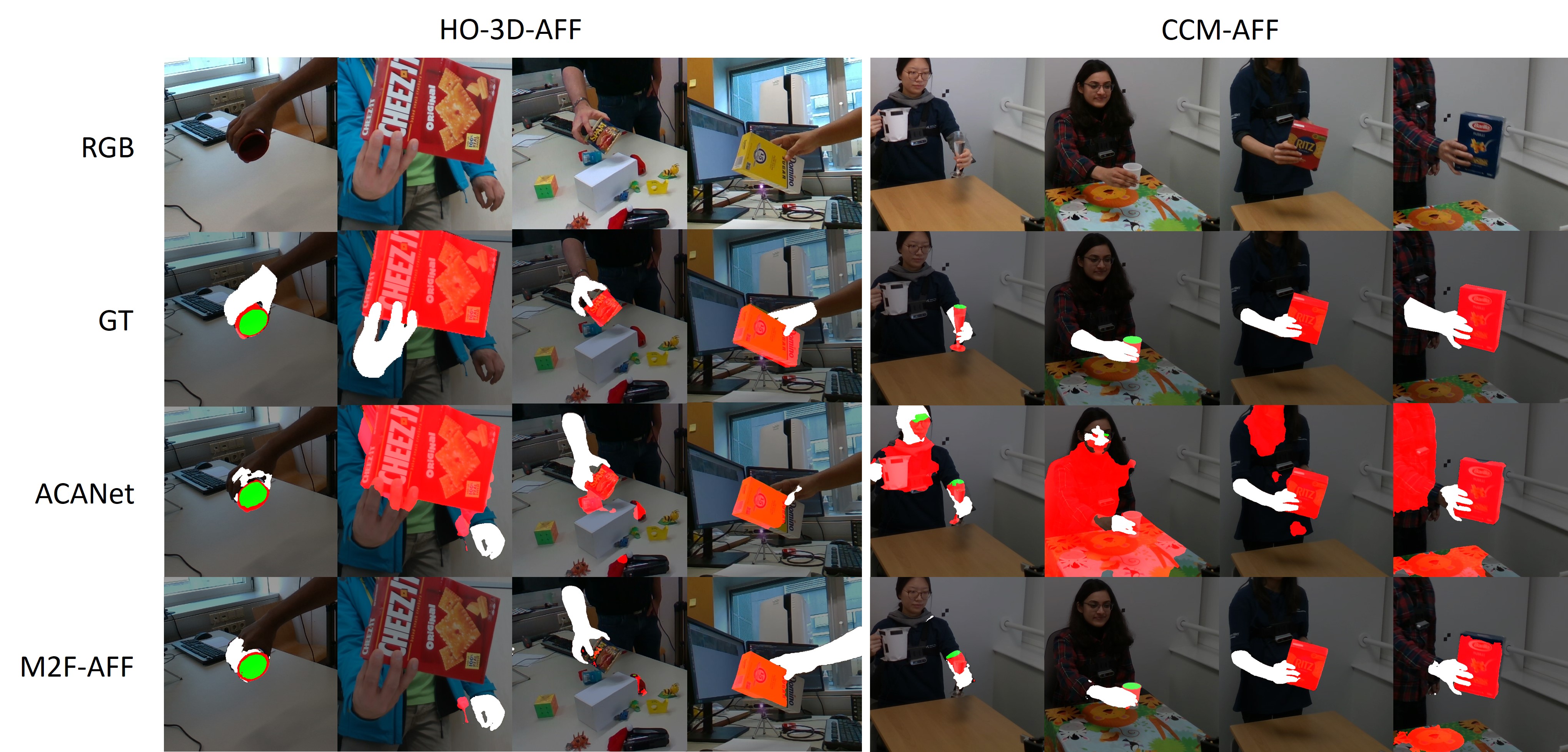}
    \caption{Comparison between the ground truth, ACANet and M2F-AFF predictions on the testing sets of HO-3D-AFF and CCM-AFF. Legend:  \protect\raisebox{2pt}{\protect\tikz \protect\draw[tgraspable,line width=2] (0,0) -- (0.3,0);}~\textit{graspable},
    \protect\raisebox{2pt}{\protect\tikz \protect\draw[tcontain,line width=2] (0,0) -- (0.3,0);}~\textit{contain},
    \protect\raisebox{1pt}{\protect\tikz \protect\draw[pattern color=black] (0,0) rectangle (0.3,0.1);}~\textit{arm}.}
    \label{fig:qual_res_ho3d_ccm}
\end{figure}

Fig.~\ref{fig:qual_res_ho3d_ccm} visually compares the segmentation results of ACANet and M2F-AFF on images from HO-3D-AFF and CCM-AFF, selected with object and hands in challenging conditions, e.g., a mug kept downward, 
a white cup held over a colored tablecloth. M2F-AFF has a more complete \textit{arm} segmentation for a challenging hand-mug pose (1st column). M2F-AFF cannot segment the \textit{graspable} region of the red box when the object is close to the camera (2nd column). However, M2F-AFF segments a more complete \textit{graspable} and \textit{arm} region than ACANet for the yellow and white box when farther from the camera (4th column). For CCM-AFF samples,  M2F-AFF predicts more complete \textit{arm} masks than ACANet. An incomplete region of the box is segmented as \textit{graspable} when there is the colored table-cloth (4th column). Compared to ACANet, M2F-AFF reduces the number of false positives on regions such as the colored tablecloth and the human clothes correctly segmenting them as background.

\section{Robustness to scale variations}

\pgfplotsset{
    boxplot/every whisker/.style={
	},
	boxplot/every box/.style={%
	},
	boxplot/every median/.style={%
	},
    boxplot/box extend=0.48,
    boxplot prepared from table/.code={
        \def\tikz@plot@handler{\pgfplotsplothandlerboxplotprepared}%
        \pgfplotsset{
            /pgfplots/boxplot prepared from table/.cd,
            #1,
        }
    },
    /pgfplots/boxplot prepared from table/.cd,
        table/.code={\pgfplotstablecopy{#1}\to\boxplot@datatable},
        row/.initial=0,
        make style readable from table/.style={
            #1/.code={
                \pgfplotstablegetelem{\pgfkeysvalueof{/pgfplots/boxplot prepared from table/row}}{##1}\of\boxplot@datatable
                \pgfplotsset{boxplot/#1/.expand once={\pgfplotsretval}}
            }
        },
        make style readable from table=lower whisker,
        make style readable from table=upper whisker,
        make style readable from table=lower quartile,
        make style readable from table=upper quartile,
        make style readable from table=median,
        make style readable from table=lower notch,
        make style readable from table=upper notch
}
\makeatother

\pgfplotstableread{data/tabletop_occupancy_percentage.txt}\whiskerumd
\pgfplotstableread{data/handoccluded_occupancy_percentage.txt}\whiskerhod
\pgfplotstableread{data/jaccard_varying_distance_umd.txt}\avgresumd
\pgfplotstableread{data/jaccard_varying_distance_handoccluded.txt}\avgreshod
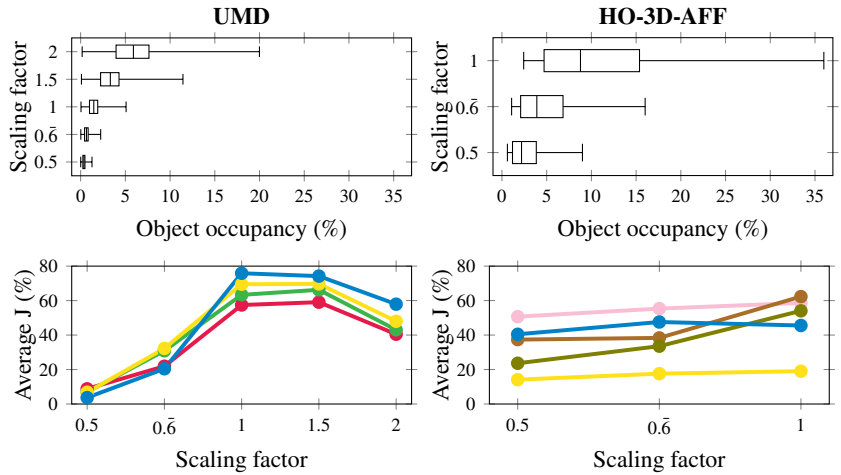
\begin{figure}[t!]
    \centering
    \begin{tikzpicture} 
	   \begin{axis}[
		boxplot/draw direction = x,
        width =0.5\linewidth,
        height=0.28\linewidth,
        xmin=-1,xmax=37,
        ymin=0.5,ymax=5.5,
	xtick = {0, 5, 10, 15, 20, 25, 30, 35},
	xticklabels = {0, 5, 10, 15, 20, 25, 30, 35},
	xlabel = {Object occupancy (\%)},
	ytick = {1, 2, 3, 4, 5},
        yticklabels = {$0.5$, $0.\bar{6}$, $1$, $1.5$, $2$},
        ylabel = {Scaling factor},
        tick label style={font=\scriptsize},
        label style={font=\footnotesize},
        ylabel style={yshift=-5pt},
        title={\textbf{UMD}},
        title style={font=\footnotesize, yshift=-4pt},
        xtick align=center, 
        ytick align=center, 
        cycle list={{black},}
	]
    \addplot+[boxplot prepared from table={
    table=\whiskerumd,
    row=4,
    lower whisker=lw,
    upper whisker=uw,
    lower quartile=lq,
    upper quartile=uq,
    median=med,
  }, boxplot prepared, draw=black 
  ]
  coordinates {};
      \addplot+[boxplot prepared from table={
    table=\whiskerumd,
    row=3,
    lower whisker=lw,
    upper whisker=uw,
    lower quartile=lq,
    upper quartile=uq,
    median=med,
  }, boxplot prepared, draw=black
  ]
  coordinates {};
      \addplot+[boxplot prepared from table={
    table=\whiskerumd,
    row=2,
    lower whisker=lw,
    upper whisker=uw,
    lower quartile=lq,
    upper quartile=uq,
    median=med
  }, boxplot prepared, draw=black
  ]
  coordinates {};
      \addplot+[boxplot prepared from table={
    table=\whiskerumd,
    row=1,
    lower whisker=lw,
    upper whisker=uw,
    lower quartile=lq,
    upper quartile=uq,
    median=med
  }, boxplot prepared, draw=black
  ]
  coordinates {};
          \addplot+[boxplot prepared from table={
    table=\whiskerumd,
    row=0,
    lower whisker=lw,
    upper whisker=uw,
    lower quartile=lq,
    upper quartile=uq,
    median=med
  }, boxplot prepared, draw=black
  ] coordinates {};
	\end{axis}
    \end{tikzpicture}
     \begin{tikzpicture} 
	   \begin{axis}[
		boxplot/draw direction = x,
        width = 0.5\linewidth,
        height=0.28\linewidth,
        xmin=-1,xmax=37,
        ymin=0.5,ymax=3.5,
	xtick = {0, 5, 10, 15, 20, 25, 30, 35},
	xlabel = {Object occupancy (\%)},
	ytick = {1, 2, 3},
        yticklabels = {$0.5$, $0.\bar{6}$, $1$},
        ylabel = {Scaling factor},
        tick label style={font=\scriptsize},
        label style={font=\footnotesize},
        ylabel style={yshift=-5pt},
        xtick align=center, 
        ytick align=center, 
        title={\textbf{HO-3D-AFF}},
        title style={font=\footnotesize, yshift=-4pt},
        cycle list={{black}}
	]
      \addplot+[boxplot prepared from table={
    table=\whiskerhod,
    row=5,
    lower whisker=lw,
    upper whisker=uw,
    lower quartile=lq,
    upper quartile=uq,
    median=med
  }, boxplot prepared, draw=black
  ]
  coordinates {};
      \addplot+[boxplot prepared from table={
    table=\whiskerhod,
    row=4,
    lower whisker=lw,
    upper whisker=uw,
    lower quartile=lq,
    upper quartile=uq,
    median=med
  }, boxplot prepared, draw=black
  ]
  coordinates {};
          \addplot+[boxplot prepared from table={
    table=\whiskerhod,
    row=3,
    lower whisker=lw,
    upper whisker=uw,
    lower quartile=lq,
    upper quartile=uq,
    median=med
  }, boxplot prepared, draw=black
  ] coordinates {};
	\end{axis}
    \end{tikzpicture}
    \begin{tikzpicture}[every axis plot/.style={line width=1.5pt, mark=*, mark size=1.8pt},]
	   \begin{axis}[
        width =0.5\linewidth,
        height=0.28\linewidth,
        xmin=0.8,xmax=5.2,
        ymin=0,ymax=80,
		xtick = {1, 2, 3, 4, 5},
        xticklabels = {$0.5$, $0.\bar{6}$, $1$, $1.5$, $2$},
		xlabel = {Scaling factor},
        ytick = {0, 20, 40, 60, 80},
        yticklabels = {0, 20, 40, 60, 80},
        ylabel = {Average J (\%)},
        ylabel style={yshift=-5pt},
        tick label style={font=\scriptsize},
        label style={font=\footnotesize},
        xtick align=center, 
        ytick align=center, 
        ]
    \addplot+ [sharp plot, dc1, mark options={fill=dc1}] table[x=CoID,y=CNN]{\avgresumd};
    \addplot+ [sharp plot, dc2, mark options={fill=dc2}] table[x=CoID,y=AffNet]{\avgresumd};
    \addplot+ [sharp plot, dc3, mark options={fill=dc3}] table[x=CoID,y=DRNAtt]{\avgresumd};
    \addplot+ [sharp plot, dc4, mark options={fill=dc4}] table[x=CoID,y=MtwoF]{\avgresumd};
    \end{axis}
    \end{tikzpicture}
    \begin{tikzpicture}[every axis plot/.style={line width=1.5pt, mark=*, mark size=1.8pt},]
    \begin{axis}[
        width =0.5\linewidth,
        height=0.28\linewidth,
        xmin=0.8,xmax=3.2,
        ymin=0,ymax=80,
	xtick = {1, 2, 3},
        xticklabels = {$0.5$, $0.\bar{6}$, $1$}, 
	xlabel = {Scaling factor},
        ytick = {0, 20, 40, 60, 80},
        yticklabels = {0, 20, 40, 60, 80},
        ylabel = {Average J (\%)},
        ylabel style={yshift=-5pt},
        tick label style={font=\scriptsize},
        label style={font=\footnotesize},
        xtick align=center, 
        ytick align=center, 
        ]
    
    \addplot+[sharp plot, dc3, mark options={fill=dc3}] table[x=CoID,y=DRNAtt]{\avgreshod};
    \addplot+[sharp plot, dc10, mark options={fill=dc10}] table[x=CoID,y=RN18U]{\avgreshod};
    \addplot+[sharp plot, dc13, mark options={fill=dc13}] table[x=CoID,y=ACANet]{\avgreshod};
    \addplot+[sharp plot, dc17, mark options={fill=dc17}] table[x=CoID,y=ACANet50]{\avgreshod};
    \addplot+ [sharp plot, dc4, mark options={fill=dc4}] table[x=CoID,y=MtwoF]{\avgreshod};
    \end{axis}
    \end{tikzpicture}
    \caption{Statistics of object pixels varying occupancy in UMD (left) and HO-3D-AFF (right) testing sets, and relative Average Jaccard Index ($J$). The whiskers plot highlights the minimum and maximum values as well as the quartiles of the occupancy distribution in each testing set. When the scaling factor is greater than $1$, images are center cropped, when it is lower than $1$, images are zero-padded. Legend:   \protect\raisebox{2pt}{\protect\tikz \protect\draw[dc1,line width=2] (0,0) -- (0.3,0);}~\textit{CNN}~\cite{nguyen2016detecting}, \protect\raisebox{2pt}{\protect\tikz \protect\draw[dc2,line width=2] (0,0) -- (0.3,0);}~\textit{AffordanceNet}~\cite{do2018affordancenet}, \protect\raisebox{2pt}{\protect\tikz \protect\draw[dc3,line width=2] (0,0) -- (0.3,0);}~\textit{DRNAtt}~\cite{gu2021visual}, 
    \protect\raisebox{2pt}{\protect\tikz \protect\draw[dc10,line width=2] (0,0) -- (0.3,0);}~\textit{RN18-U}, 
    \protect\raisebox{2pt}{\protect\tikz \protect\draw[dc13,line width=2] (0,0) -- (0.3,0);}~\textit{ACANet}~\cite{apicella2023affordance},
    \protect\raisebox{2pt}{\protect\tikz \protect\draw[dc17,line width=2] (0,0) -- (0.3,0);}~\textit{ACANet50}~\cite{apicella2023affordance}, \protect\raisebox{2pt}{\protect\tikz \protect\draw[dc4,line width=2] (0,0) -- (0.3,0);}~\textit{M2F-AFF}.}
    \label{fig:varying_scale_results}
 \end{figure}

To analyse the performance of models when varying object scale, we simulate a zoom-out in the images dividing height and width by $1.5$, $2$, and zero-padding images to respect the input resolution of models. A zoom-in effect is applied by multiplying height and width by $1.5$, $2$ and then center cropping the images to the model input resolution. We assess the performance in UMD for the tabletop setting and HO-3D-AFF for the hand-occluded setting. In UMD the object is framed in the image center and the distance with the camera is fixed, allowing for both scale variations without losing image quality. HO-3D-AFF is the testing set with the highest object scale (occupancy with a maximum over $35\%$, see Supp. Mat.) and where  has average performance lower than other methods. Therefore, we zoom-out the images to shift the distribution closer to the one of other datasets with hand-occluded objects. 

Fig.~\ref{fig:varying_scale_results} (top left) shows the statistics of object occupancy in UMD testing set. The original testing set of UMD has a low statistic of the closest object occupancy being approximately $5\%$. As expected, by zooming-out, the statistics is shifted towards values under $5\%$ and conversely the zooming-in causes the statistics to increase. Fig.~\ref{fig:varying_scale_results} (bottom left) shows the average Jaccard Index of re-implemented methods on the UMD testing sets when varying object occupancy. Methods achieve a higher average Jaccard Index in zoomed-in testing sets than zoomed-out, due to the fact that the scaling in $[1, 1.5]$ is present in the training augmentation. M2F-AFF outperforms other methods when the testing set images are zoomed-in. When images are zoomed-out, the best performing models are AffNet and CNN.
Fig.~\ref{fig:varying_scale_results} (top right) shows the object occupancy of images in HO-3D-AFF, progressively shifting to lower values when zooming-out images to reduce object scale. As shown in Fig.~\ref{fig:varying_scale_results} (bottom right), RN18-U outperforms all models when the scaling factor is lower than 1, while M2F-AFF is the second best-performing model. 
Both ACANet and ACANet50 have an increase of more than $20$ percentage points from the scaling factor of $0.\bar{6}$ to $1$. DRNAtt is the only model that keeps low generalisation, not only to HO-3D-AFF testing set but also to scale variation. These results suggest that the generalisation to different object scales remains an open challenge for affordance segmentation, providing a potential direction for future contributions.

\begin{figure}[t!]
    \centering
    \includegraphics[width=\linewidth]{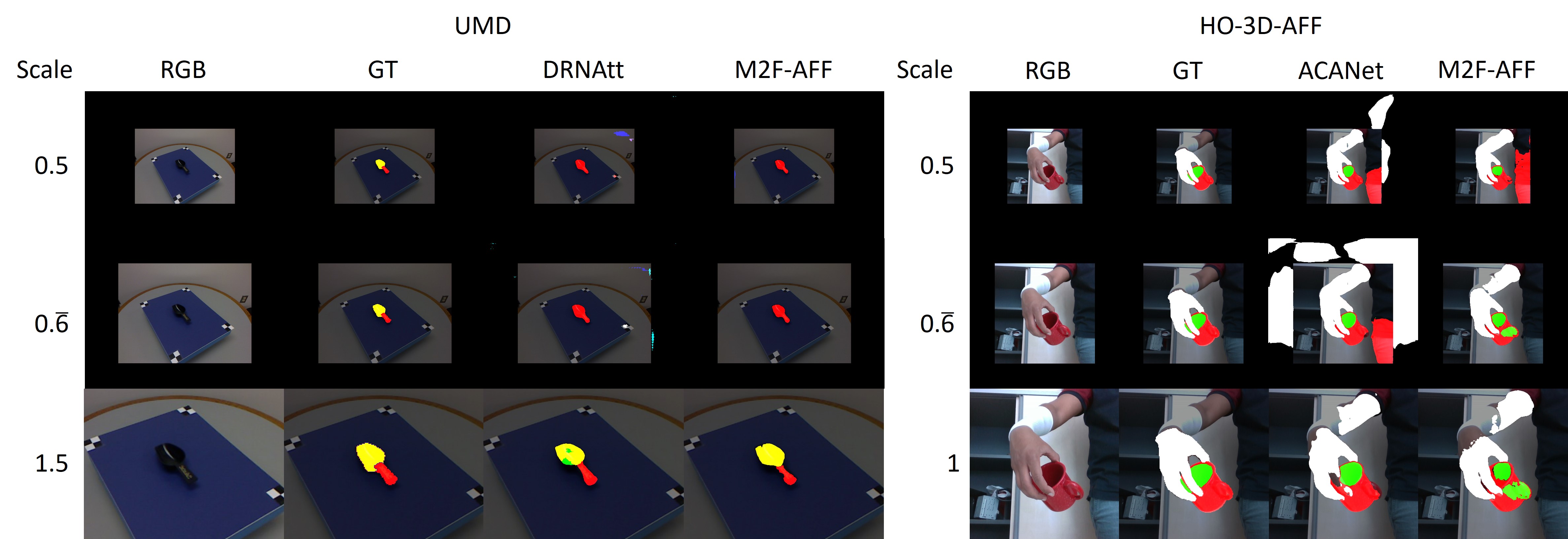}
    \caption{Comparison between the segmentation ground-truths, and predictions of DRNAtt~\cite{gu2021visual}, ACANet~\cite{apicella2023affordance}, M2F-AFF, on UMD and CCM-AFF testing sets varying object scale. Legend:
    \protect\raisebox{2pt}{\protect\tikz \protect\draw[tgraspable,line width=2] (0,0) -- (0.3,0);}~\textit{graspable},
    \protect\raisebox{2pt}{\protect\tikz \protect\draw[tcut,line width=2] (0,0) -- (0.3,0);}~\textit{cut},
    \protect\raisebox{2pt}{\protect\tikz \protect\draw[tscoop,line width=2] (0,0) -- (0.3,0);}~\textit{scoop},
    \protect\raisebox{2pt}{\protect\tikz \protect\draw[tcontain,line width=2] (0,0) -- (0.3,0);}~\textit{contain},
    \protect\raisebox{2pt}{\protect\tikz \protect\draw[tpound,line width=2] (0,0) -- (0.3,0);}~\textit{pound},
    \protect\raisebox{2pt}{\protect\tikz \protect\draw[tsupport,line width=2] (0,0) -- (0.3,0);}~\textit{support},
    \protect\raisebox{2pt}{\protect\tikz \protect\draw[twgrasp,line width=2] (0,0) -- (0.3,0);}~\textit{wrap-grasp},
    \protect\raisebox{1pt}{\protect\tikz \protect\draw[pattern color=black] (0,0) rectangle (0.3,0.1);}~\textit{arm}.}
    \label{fig:qual_res_varying_scale}
\end{figure}

Fig.~\ref{fig:qual_res_varying_scale} shows the comparison between models predictions when varying object scale. In UMD, when objects are scaled down, both DRNAtt and M2F-AFF produce false positives also outside the object region (first two rows), when the objects are large scaled up, the false positive are inside the object region (DRNAtt last row). In HO-3D-AFF, when objects are scaled down, ACANet produces false positives for the \textit{arm} class in the additional zero-padding and for the \textit{graspable} class on the human body, while M2F-AFF has similar predictions and produces false positives for \textit{graspable} on the human body with the lowest object scale.

\section{Conclusion}
\label{sec:conclusions}
We identified reproducibility issues with affordance segmentation methods. These issues range from unknown training setups to unfair comparative evaluations. 
We re-implemented and evaluated a range of selected methods under a uniform training setup and common testing sets across two datasets. By re-training  Mask2Former, a recent architecture based on latent vectors that was not yet considered for affordance segmentation, we showed that this model outperforms other methods on the testing set with objects on a tabletop and on most testing sets in the hand-occluded settings. Moreover, we analysed and showed how the methods are not robust to scale variations despite the use of scaling in the data augmentation during training. As future work, we will extend the benchmarking  evaluation to datasets with multiple objects in an image and where occlusions are caused by other objects.


\bibliographystyle{eccv_2024/splncs04}
\bibliography{refs}

\definecolor{dc1}{RGB}{230, 25, 75} 
\definecolor{dc2}{RGB}{60, 180, 75} 
\definecolor{dc3}{RGB}{255, 225, 25} 
\definecolor{dc4}{RGB}{0, 130, 200} 
\definecolor{dc5}{RGB}{245, 130, 48} 
\definecolor{dc6}{RGB}{145, 30, 180} 
\definecolor{dc7}{RGB}{70, 240, 240} 
\definecolor{dc8}{RGB}{240, 50, 230} 
\definecolor{dc9}{RGB}{210, 245, 60} 
\definecolor{dc10}{RGB}{250, 190, 212} 
\definecolor{dc11}{RGB}{0, 128, 128} 
\definecolor{dc12}{RGB}{220, 190, 255} 
\definecolor{dc13}{RGB}{170, 110, 40} 
\definecolor{dc14}{RGB}{255, 250, 200} 
\definecolor{dc15}{RGB}{128, 0, 0} 
\definecolor{dc16}{RGB}{170, 255, 195} 
\definecolor{dc17}{RGB}{128, 128, 0} 
\definecolor{dc18}{RGB}{255, 215, 180} 
\definecolor{dc19}{RGB}{0, 0, 128} 
\definecolor{dc20}{RGB}{128, 128, 128} 
\definecolor{dc21}{RGB}{255, 255, 255} 
\definecolor{dc22}{RGB}{0, 0, 0} 

\definecolor{mylightgray}{gray}{0.9}
\definecolor{tgraspable}{RGB}{255, 0, 0}
\definecolor{tcut}{RGB}{0, 0, 255}
\definecolor{tscoop}{RGB}{255, 255, 0}
\definecolor{tcontain}{RGB}{0, 255, 0}
\definecolor{tpound}{RGB}{128, 80, 215}
\definecolor{tsupport}{RGB}{0, 255, 255}
\definecolor{twgrasp}{RGB}{196, 164, 132}
\definecolor{tarm}{RGB}{255, 255, 255}


\title{Supplementary Material document \\
Segmenting Object Affordances:\\ Reproducibility and Sensitivity to Scale} 

\titlerunning{Supplementary: Segmenting Object Affordances}

\author{Tommaso Apicella\inst{1,2}\orcidlink{0000-0001-9001-5641} \and
Alessio Xompero\inst{2}\orcidlink{0000-0002-8227-8529} \and
Paolo Gastaldo\inst{1}\orcidlink{0000-0002-5748-3942
}\and \\
Andrea Cavallaro\inst{2,3}\orcidlink{0000-0001-5086-7858}}

\authorrunning{T.~Apicella et al.}

\institute{University of Genoa, Italy \and
Queen Mary University of London, United Kingdom
 \and
Idiap Research Institute, Switzerland \and 
\'{E}cole Polytechnique F\'{e}d\'{e}rale de Lausanne, Switzerland\\
\email{\{t.apicella, a.xompero\}@qmul.ac.uk} \\ 
\email{paolo.gastaldo@unige.it}\\  
\email{a.cavallaro@idiap.ch}}

\maketitle


\noindent We provide additional details about the performance measures and additional analyses about affordance segmentation results. 

\section{Performance measures}
Given the set of images with the associated annotations $\{I_i, M_i\}^{N}_{i=1}$, $I \in \mathbb{R}^{W \times H \times 3}$, $M \in \{0, ..., S-1 \}^{W \times H}$, $S$ is the number of segmentation classes, $W$ is width, $H$ is height, the prediction of the model is the set $\{\hat{M}_i\}^{N}_{i=1}$ with $\hat{M}_i \in \{ 0, ..., S-1 \}^{W \times H}$. For each segmentation class, we consider the annotated and the predicted segmentation mask with the column-stack representation $Y$ and $\hat{Y}$ respectively. 

We consider a pixel $x$ having value $s$ in the prediction $\hat{Y}$ and in the annotation $Y$ as true positive ($TP$), a pixel having value $s$ in $\hat{Y}$, but not in $Y$ as false positive ($FP$), a pixel having value $s$ in $Y$, but not in $\hat{Y}$ as false negative ($FN$). 
Per-class precision ($P$) measures how many pixels predicted as class $s$ are correct, among all pixels predicted as class $s$:  
\begin{equation}
    P = \frac{\sum^{N}_{i=1} \sum^{W \times H}_{x} TP}{\sum^{N}_{i=1} \sum^{W \times H}_{x} TP + FP} \: .
\end{equation} 
Per-class recall ($R$) measures how many pixels predicted as class $s$ are correct, among all pixels annotated as class $s$:  
\begin{equation}
    R = \frac{\sum^{N}_{i=1} \sum^{W \times H}_{x} TP}{\sum^{N}_{i=1} \sum^{W \times H}_{x} TP + FN} \: .
\end{equation}
Per-class Jaccard index ($J$) measures how many pixels predicted as class $s$ are correct, among all pixels (predicted and annotated):
\begin{equation}
    J = \frac{\sum^{N}_{i=1} \sum^{W \times H}_{x} TP}{\sum^{N}_{i=1} \sum^{W \times H}_{x} TP + FP + FN} \: .
\end{equation}

The weighting function $\mathcal{A}$ captures the dependency between foreground and background pixels
\begin{equation}
    \mathcal{A}(i, j) = \begin{cases}
                \frac{1}{\sqrt{2\pi\sigma^{2}}} e^{-\frac{d(i,j)^{2}}{2\sigma^{2}}} & \forall i, j\quad Y(i)=1, Y(j)=1 \\
                    1& \forall i, j\quad Y(i)=0, i=j \\
                    0& \text{otherwise}
                \end{cases} \:,
\end{equation}
where  $d(i,j)$ is the Euclidean distance between pixel $i$ and pixel $j$, $\sigma^2$ controls the influence of pixels that are farther away. The larger $\sigma^2$ is, the greater the influence of distant pixels.
    
The function $D$ represents the varying importance of pixels based on their distance from the foreground:
\begin{equation}
        D(i)= \begin{cases}
                1 & \forall i, Y(i)=1 \\
                2-e^{\alpha\cdot\Delta(i)} & \text{otherwise}
              \end{cases} \:,
\end{equation}
where $\Delta(i) = \min_{Y(j) = 1} d(i,j)$. The constant $\alpha$ determines the decay rate.

The weighted error is a vector with values in $[0, 1]$ range, $E^{w} = \min(E, E\mathcal{A}) \cdot {D}$ and combines the error $E = |Y - \hat{Y}|$ with the weighting functions $\mathcal{A}$ and $D$. A true positive is a pixel that has value complementary to the weighted error and belongs to the annotation $TP^{w}=(1-E^{w}) \cdot Y$. A false positive is a pixel of the weighted error that does not belongs to the annotation $FP^{w}=E^{w}\cdot(1-Y)$. A false negative is a pixel of the weighted error that belongs to the annotation and $FN^{w}=E^{w}\cdot Y$. The per-class weighted precision ($P^w$) measures the amount of correctly segmented pixels among all predicted pixels the based on the weighting functions
\begin{equation}
    P^{w} = \frac{ \sum^{N}_{i=1} \sum^{W \times H}_{x} TP^{w}}{ \sum^{N}_{i=1} \sum^{W \times H}_{x} TP^{w} + FP^{w}} \:.
\end{equation}

The per-class weighted recall ($R^w$) measures the amount of correctly segmented pixels among all annotated pixels based on the weighting functions
\begin{equation}
    R^{w} = \frac{ \sum^{N}_{i=1} \sum^{W \times H}_{x} TP^{w}}{\sum^{N}_{i=1} \sum^{W \times H}_{x}TP^{w} + FN^{w}} \:.
\end{equation}

The per-class $F^{w}_{\beta}$ between prediction and annotation is obtained combining the Precision $P^w$ and the Recall $R^w$
\begin{equation}
    F^{w}_{\beta} = (1 + \beta^2) \frac{P^{w} \cdot R^{w}}{\beta^2 \cdot P^{w} + R^{w}} \:,
\end{equation}

To evaluate the performance of affordance segmentation methods, existing works use precision, recall, and Jaccard index for the scenario with hand-occluded objects~\cite{apicella2023affordance}, and $F^{w}_{\beta}$ for the scenario with unoccluded objects~\cite{margolin2014evaluate}.

\section{Object pixels occupancy}
We report in Fig.~\ref{fig:pixel_occupancy} the object occupancy across the testing sets in pixels percentage used as proxy to objects scale: the higher the percentage, the bigger the object. Note that in UMD~\cite{myers2015affordance} all the objects occupy at most $5\%$ of the images. In the hand-occluded testing sets, the occupancy statistics is similar between CCM-AFF~\cite{xompero2020corsmal,apicella2023affordance} and the mixed-reality testing sets, while in HO3D-AFF~\cite{hampali2020honnotate,apicella2023affordance} over $25\%$ of objects occupies more pixels than other testing sets. HO-3D-AFF is the testing set with the highest object scale with maximum occupancy over $35\%$.

\pgfplotsset{
    boxplot/every whisker/.style={
	},
	boxplot/every box/.style={%
	},
	boxplot/every median/.style={%
	},
    boxplot/box extend=0.48,
    boxplot prepared from table/.code={
        \def\tikz@plot@handler{\pgfplotsplothandlerboxplotprepared}%
        \pgfplotsset{
            /pgfplots/boxplot prepared from table/.cd,
            #1,
        }
    },
    /pgfplots/boxplot prepared from table/.cd,
        table/.code={\pgfplotstablecopy{#1}\to\boxplot@datatable},
        row/.initial=0,
        make style readable from table/.style={
            #1/.code={
                \pgfplotstablegetelem{\pgfkeysvalueof{/pgfplots/boxplot prepared from table/row}}{##1}\of\boxplot@datatable
                \pgfplotsset{boxplot/#1/.expand once={\pgfplotsretval}}
            }
        },
        make style readable from table=lower whisker,
        make style readable from table=upper whisker,
        make style readable from table=lower quartile,
        make style readable from table=upper quartile,
        make style readable from table=median,
        make style readable from table=lower notch,
        make style readable from table=upper notch
}
\makeatother

\pgfplotstableread{data/handoccluded_occupancy_percentage.txt}\datatable
\pgfplotstableread{data/tabletop_occupancy_percentage.txt}\datatableumd

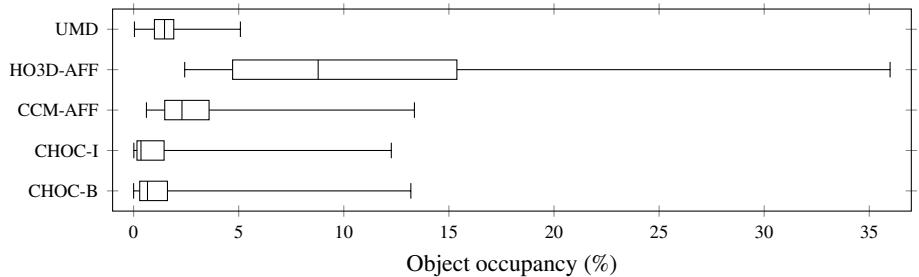
\begin{figure}[t!]
    \centering
    \begin{tikzpicture} 
	   \begin{axis}[
		boxplot/draw direction = x,
        width = \linewidth,
        height=.35\linewidth,
        xmin=-1,xmax=37,
        ymin=0.5,ymax=5.5,
		xtick = {0, 5, 10, 15, 20, 25, 30, 35},
		xticklabels = {0, 5, 10, 15, 20, 25, 30, 35},
		xlabel = {Object occupancy (\%)},
		ytick = {1, 2, 3, 4, 5}, 
        yticklabels = {CHOC-B, CHOC-I, CCM-AFF, HO3D-AFF, UMD}, 
        tick label style={font=\scriptsize},
        xtick align=center, 
        ytick align=center, 
        cycle list={{black}}
	]
    \addplot+[boxplot prepared from table={
    table=\datatable,
    row=0,
    lower whisker=lw,
    upper whisker=uw,
    lower quartile=lq,
    upper quartile=uq,
    median=med,
  }, boxplot prepared, draw=black 
  ]
  coordinates {};
      \addplot+[boxplot prepared from table={
    table=\datatable,
    row=1,
    lower whisker=lw,
    upper whisker=uw,
    lower quartile=lq,
    upper quartile=uq,
    median=med,
  }, boxplot prepared, draw=black
  ]
  coordinates {};
      \addplot+[boxplot prepared from table={
    table=\datatable,
    row=2,
    lower whisker=lw,
    upper whisker=uw,
    lower quartile=lq,
    upper quartile=uq,
    median=med
  }, boxplot prepared, draw=black
  ]
  coordinates {};
      \addplot+[boxplot prepared from table={
    table=\datatable,
    row=3,
    lower whisker=lw,
    upper whisker=uw,
    lower quartile=lq,
    upper quartile=uq,
    median=med
  }, boxplot prepared, draw=black
  ]
  coordinates {};
  \addplot+[boxplot prepared from table={
    table=\datatableumd,
    row=2,
    lower whisker=lw,
    upper whisker=uw,
    lower quartile=lq,
    upper quartile=uq,
    median=med
  }, boxplot prepared, draw=black
  ]
  coordinates {};
	\end{axis}
    \end{tikzpicture}
    \caption{Statistics of object pixels occupancy in unoccluded and hand-occluded testing sets. }
    \label{fig:pixel_occupancy}
\end{figure}

\section{Quantitative results}
\begin{table*}[t!]
    \centering
    \caption{Comparison of affordance segmentation methods using $F^w_\beta$ on UMD testing set. The first group results are taken from the papers, while in the second group methods are re-implemented and trained using our setup, common to all models.}
    \begin{tabular}{@{}rr c c c c c c c c @{}}
    \toprule
    Training setup & {Model} & \multicolumn{1}{c}{\textit{grasp}} & \multicolumn{1}{c}{\textit{cut}} & \multicolumn{1}{c}{\textit{scoop}} & \multicolumn{1}{c}{\textit{contain}} & \multicolumn{1}{c}{\textit{pound}} & \multicolumn{1}{c}{\textit{support}} & \multicolumn{1}{c}{\textit{wrap-grasp}} & \multicolumn{1}{c}{avg} \\
    \midrule
    \multirow{3}{*}{Literature}
    & CNN~\cite{nguyen2016detecting} & 71.90 & 73.70 & 74.40 & 81.70 & 79.40 & 78.00 & 76.90 & 76.57\\
    & AffNet~\cite{do2018affordancenet} & 73.10 & 76.20 & 79.30 & 83.30 & 83.60 & 82.10 & 81.40 & 79.85\\
    & DRNAtt~\cite{gu2021visual} & 90.40 & 92.40 & 92.70 & 95.40 & 95.60 & 96.30 & 95.50 & 94.04 \\         
    \midrule 
    \multirow{3}{*}{Ours} 
    & CNN~\cite{nguyen2016detecting} & 62.38 & 77.36 & 39.77 & 85.03 & 63.38 & 53.77 & 78.47 & 65.74 \\
    & AffNet~\cite{do2018affordancenet} & 66.95 & 80.46 & 43.60 & 89.96 & 67.17 & 80.76 & 91.69 & 69.83 \\
    & DRNAtt~\cite{gu2021visual} & 61.50  & 80.10 & 41.51 & 89.98 & 70.61 & 54.11 & 91.03 & 74.37 \\                
    \bottomrule 
    \addlinespace[\belowrulesep]
   \multicolumn{10}{l}{\parbox{0.95\columnwidth}{\scriptsize{AffNet:AffordanceNet}}}
    \end{tabular}
    \label{tab:res_umd_pre_post}
\end{table*}
In Table~\ref{tab:res_umd_pre_post}, we compare the $F^w_\beta$ performance of models in the respective papers and the performance of models re-implemented and re-trained using our setup. The trends between both setups are similar and the ranking is the same, however the values obtained with our setup are lower than the ones in the literature. Our training setup is shared among models, hence it is more fair than the evaluation provided in the previous works. DRNAtt is the model changing the most the performance values between the two setups with a difference of almost $20$ percentage points. 

In Table~\ref{tab:tableresults_handocclusions}, we report the performance of models on hand-occluded testing sets using Precision (P), Recall (R) and Jaccard Index (J). Results between methods are similar in mixed-reality testing sets CHOC-I and CHOC-B. ACANet is the best performing model on HO-3D-AFF. On CCM-AFF, M2F-AFF improves the precision in all classes due predicting less false positives on the table and on the human body than other methods. M2F-AFF has also higher recall compared to other models in both \textit{contain} and \textit{arm} classes.

\begin{table}[t!]
    \centering
    \caption{Comparison of the affordance and arm segmentation results between the models on the two mixed-reality testing sets and on the two real testing sets.}
    \begin{tabular}{@{}lr rrr rrr rrr r @{}}
    \toprule
    Testing set & {Model} & \multicolumn{3}{c}{\textit{graspable}} & \multicolumn{3}{c}{\textit{contain}} & \multicolumn{3}{c}{\textit{arm}} & \multicolumn{1}{c}{\textit{overall}} \\
    \cmidrule(lr){3-5}\cmidrule(lr){6-8}\cmidrule(lr){9-11}
    & & \multicolumn{1}{c}{$P$} & \multicolumn{1}{c}{$R$} & \multicolumn{1}{c}{$J$} & \multicolumn{1}{c}{$P$} & \multicolumn{1}{c}{$R$} & \multicolumn{1}{c}{$J$} & \multicolumn{1}{c}{$P$} & \multicolumn{1}{c}{$R$} & \multicolumn{1}{c}{$J$} & \multicolumn{1}{c}{$J$} \\
    \midrule
    \multirow{6}{*}{CHOC-B} 
    & RN50-F & 97.33 & 95.72 & 93.27 & 89.83 & 91.94 & 83.27 & - & - & - & - \\
    & RN18-U & 96.79 & 96.44 & 93.45 & 84.94 & 93.16 & 79.95 & 96.55 & 96.46 & 93.24 & 88.88 \\
    & DRNAtt & 96.38 & 97.04 & 93.63 & 91.84 & 90.63 & 83.88 & 96.94 & 97.19 & 94.30 & 90.60 \\
    & ACANet & 97.09 & 96.60 & 93.88 & 89.46 & 94.67 & 85.17 & 96.48 & 96.52 & 93.24 & 90.76 \\
    & ACANet50 & 97.17 & 96.64 & 94.00 & 89.85 & 94.73 & 85.57 & 96.59 & 96.91 & 93.70 & 91.09 \\ 
    & M2F-AFF & \textbf{98.05} & \textbf{97.33} & \textbf{95.48} & \textbf{92.81} & \textbf{95.14} & \textbf{88.61} & \textbf{97.81} & \textbf{97.43} & \textbf{95.36} & \textbf{93.15} \\ 
    \midrule
    \multirow{6}{*}{CHOC-I}
     & RN50-F & 96.55 & 95.35 & 92.20 & 90.20 & 74.27 & 68.73 & - & - & - & - \\
     & RN18-U & 96.33 & 96.35 & 92.94 & 88.97 & 74.32 & 68.04 & 96.67 & 96.91 & 93.78 & 84.92 \\
    & DRNAtt & 95.85 & 96.74 & 92.85 & 90.48 & 71.08 & 66.13 & 97.00 & 96.88 & 94.07 & 84.35 \\
    & ACANet & 96.36 & 96.51 & 93.11 & 88.72 & 76.68 & 69.86 & 96.94 & 96.77 & 93.90 & 85.62 \\
    & ANet50 & 96.77 & 96.37 & 93.37 & 89.29 & 79.60 & 72.66 & 96.88 & 97.01 & 94.07 &  86.70 \\ 
    & M2F-AFF & \textbf{97.51} & \textbf{97.63} & \textbf{95.26} & \textbf{91.17} & \textbf{83.93} & \textbf{77.62} & \textbf{98.00} & \textbf{97.97} & \textbf{96.04} &  \textbf{89.64} \\ 
    \midrule
    \multirow{6}{*}{HO-3D-AFF} 
     & RN50-F & \textbf{95.61} & 18.29 & 18.14 & \textbf{90.69} & 79.57 & 73.56 & - & - & - & - \\
     & RN18-U & 85.85 & 72.53 & 64.79 & 88.21 & 87.61 & \textbf{78.42} & 61.80 & 41.03 & 32.73 & 58.64\\
     & DRNAtt & 75.42 & 44.08 & 38.54 & 87.26 & 18.75 & 18.25 & 50.23 & 0.32 & 0.32 & 19.04 \\
     & ACANet & 89.72 & \textbf{80.78} & \textbf{73.93} & 79.20 & \textbf{90.43} & 73.07 & \textbf{61.95} & \textbf{53.02} & \textbf{40.00} & \textbf{62.33} \\
     & ACANet50 & 9.88 & 78.05 & 58.40 & 75.72 & 81.20 & 64.43 & 64.42 & 50.29 & 39.36 & 54.06 \\ 
     & M2F-AFF & 87.69 & 39.41 & 37.35 & 76.59 & 81.48 & 65.24 & 58.00 & 45.28 & 34.10 &  45.56 \\ 
    \midrule
    \multirow{6}{*}{CCM-AFF} 
     & RN50-F & 6.14 & 87.87 & 6.09 & 13.51 & 33.11 & 10.61 & - & - & - & - \\
     & RN18-U & 13.69 & 78.69 & 13.20 & 31.92 & 42.44 & 22.28 & 44.21 & 42.53 & 27.68 & 21.05 \\
     & DRNAtt & 6.37 & \textbf{95.09} & 6.35 & 0.00 & 0.00 & 0.00 & 4.47 & 0.24 & 0.23 & 2.19 \\
     & ACANet & 10.22 & 86.50 & 10.06 & 45.40 & 37.46 & 25.83 & 49.47 & 45.35 & 31.00 & 22.30 \\
     & ACANet50 & 8.29 & 80.03 & 8.12 & 69.88 & 18.85 & 17.43 & 61.23 & 40.98 & 32.54 & 19.36 \\ 
     & M2F-AFF & \textbf{36.99} & 63.44 & \textbf{30.49} & \textbf{69.54} & \textbf{54.92} & \textbf{44.27} & \textbf{70.61} & \textbf{68.54} & \textbf{53.32} & \textbf{42.69} \\ 
    \bottomrule \addlinespace[\belowrulesep]
    \multicolumn{12}{l}{\parbox{0.95\columnwidth}{\scriptsize{Highlighted in \textbf{bold} the best performing results. KEY -- P: per-class Precision, R: per-class Recall, J: per-class Jaccard index, RN50-F:~ResNet50-FastFCN~\cite{hussain2020fpha}, RN18-U:~ResNet18-UNET, DRNAtt~\cite{gu2021visual}, ACANet~\cite{apicella2023affordance}, ACANet50: ACANet with ResNet50 backbone, M2F-AFF: Mask2Former~\cite{cheng2022masked} trained to segment affordances. CHOC-B:~the CHOC-AFF testing set with new backgrounds, CHOC-I:~the CHOC-AFF testing set with new instances.}}}
    \end{tabular}
    \label{tab:tableresults_handocclusions}
\end{table}

\section{Segmentation predictions}
To complement the segmentation results analysis of the main paper, we show additional predictions of models. 

\begin{figure}[t!]
    \centering
    \includegraphics[width=1\linewidth]{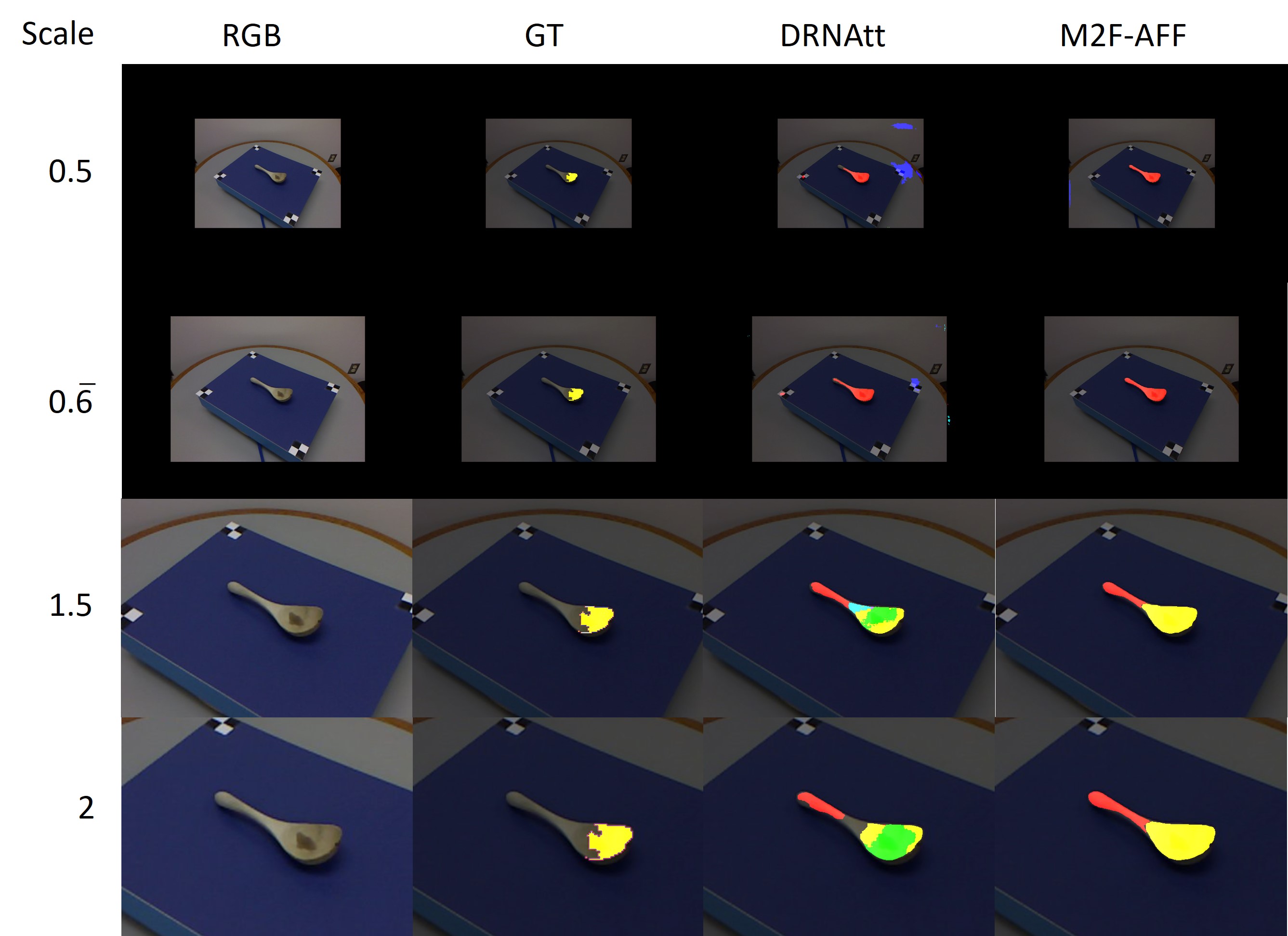}
    \includegraphics[width=1\linewidth]{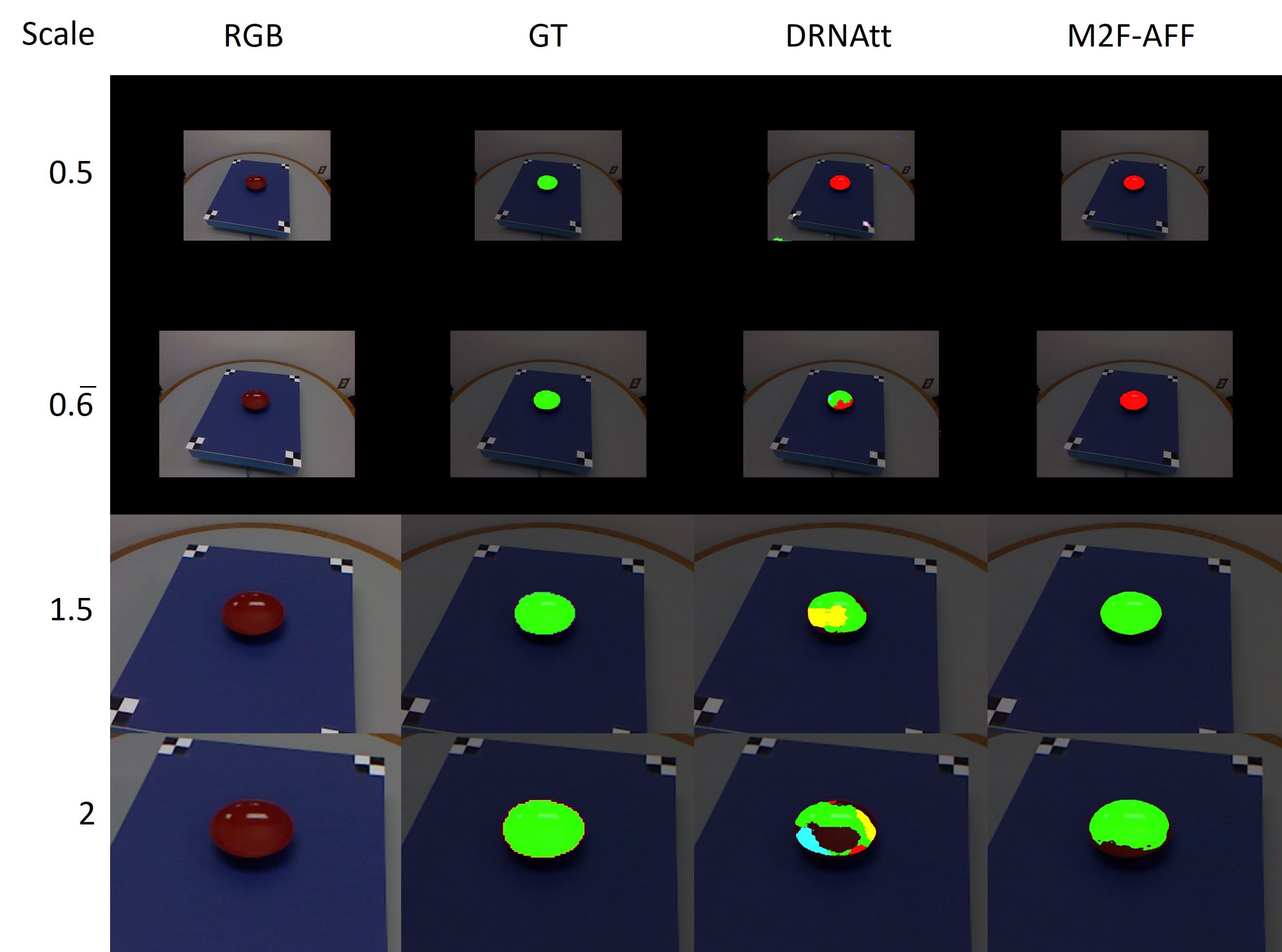}
    \caption{Comparison between the segmentation annotation, and predictions of DRNAtt~\cite{gu2021visual} and M2F-AFF on UMD testing set varying object scale. Legend:
    \protect\raisebox{2pt}{\protect\tikz \protect\draw[tgraspable,line width=2] (0,0) -- (0.3,0);}~\textit{graspable},
    \protect\raisebox{2pt}{\protect\tikz \protect\draw[tcut,line width=2] (0,0) -- (0.3,0);}~\textit{cut},
    \protect\raisebox{2pt}{\protect\tikz \protect\draw[tscoop,line width=2] (0,0) -- (0.3,0);}~\textit{scoop},
    \protect\raisebox{2pt}{\protect\tikz \protect\draw[tcontain,line width=2] (0,0) -- (0.3,0);}~\textit{contain},
    \protect\raisebox{2pt}{\protect\tikz \protect\draw[tpound,line width=2] (0,0) -- (0.3,0);}~\textit{pound},
    \protect\raisebox{2pt}{\protect\tikz \protect\draw[tsupport,line width=2] (0,0) -- (0.3,0);}~\textit{support},
    \protect\raisebox{2pt}{\protect\tikz \protect\draw[twgrasp,line width=2] (0,0) -- (0.3,0);}~\textit{wrap-grasp}.}
    \label{fig:qual_umd_varying_scale}
\end{figure}

Fig.~\ref{fig:qual_umd_varying_scale} compares the predictions of M2F-AFF and DRNAtt with the annotation. In the spoon case (top figure), the annotation misses the \textit{graspable} segmentation of the handle, yet both models segments correctly as \textit{graspable} the handle when object scale is higher than 1. In both spoon and bowl cases M2F-AFF has no false positives, while the segmentation of DRNAtt has segments also wrong classes on both objects. When the scaling factor is less than 1, M2F-AFF segments the wrong class but keeps a low amount of false positives outside the object region.  

\begin{figure}[t!]
    \centering
    \includegraphics[width=1\linewidth]{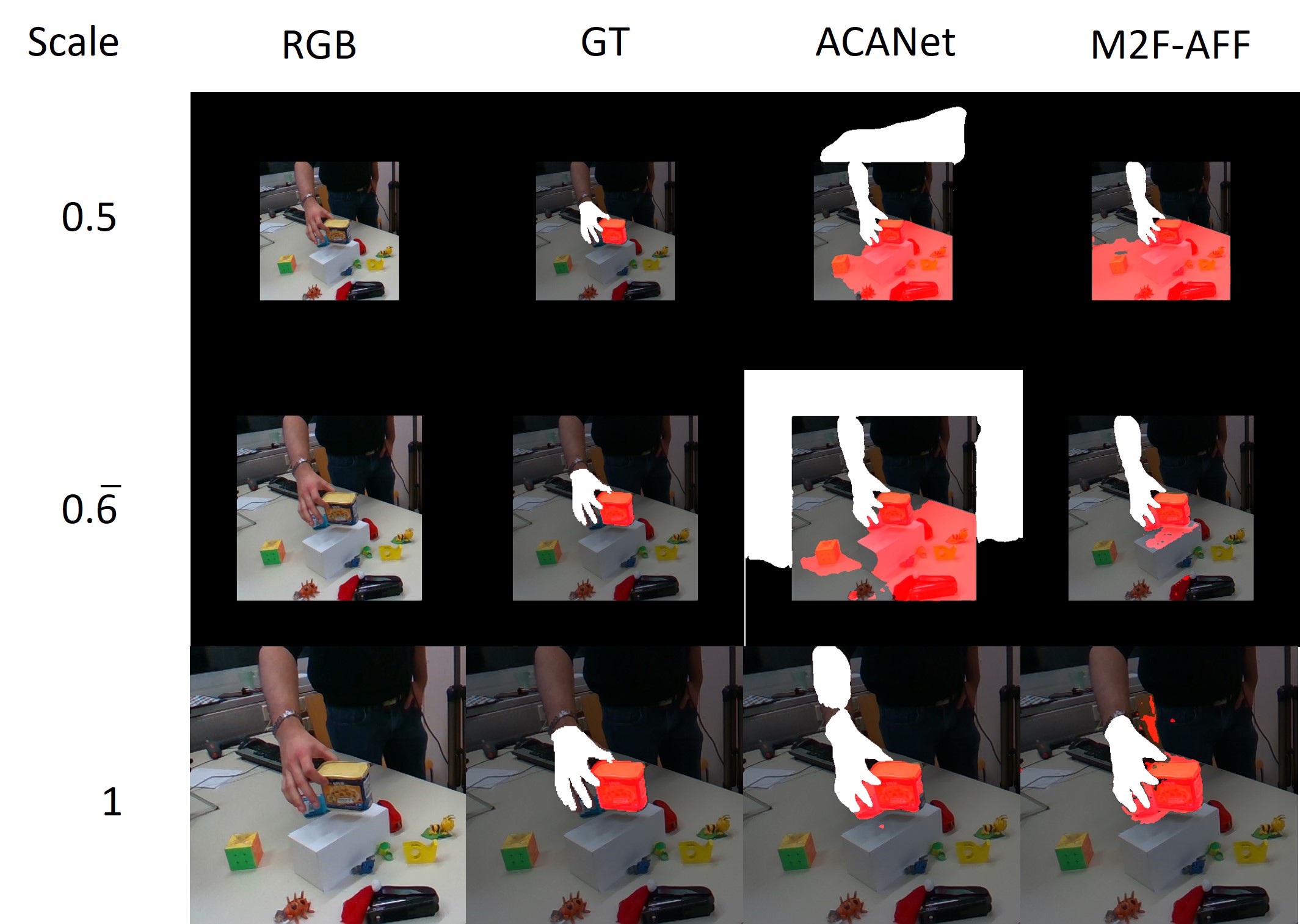}
    \includegraphics[width=1\linewidth]{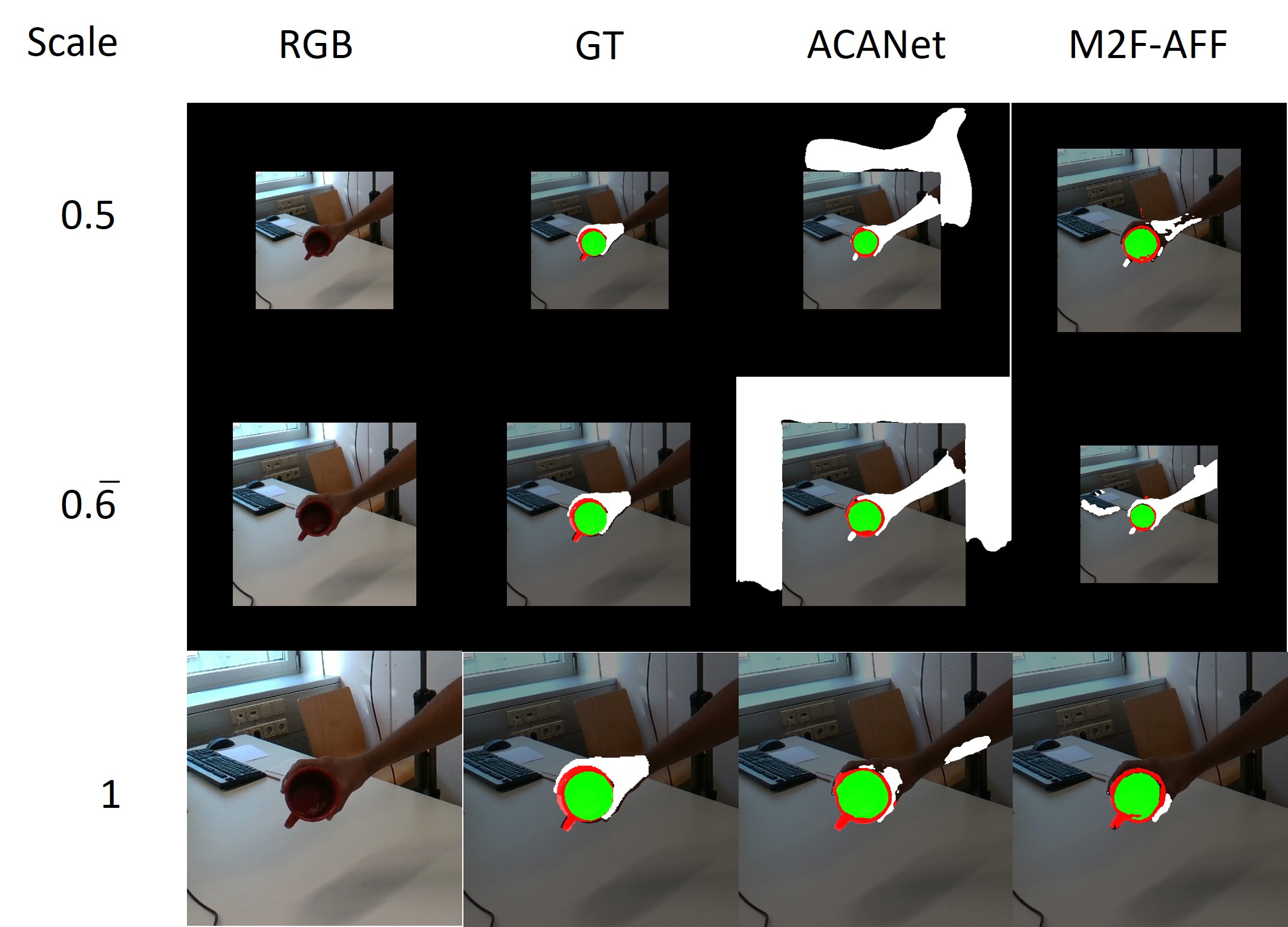}
    \caption{Comparison between the segmentation annotation, and predictions of ACANet~\cite{apicella2023affordance} and M2F-AFF on CCM-AFF testing sets varying object scale. Legend:
    \protect\raisebox{2pt}{\protect\tikz \protect\draw[tgraspable,line width=2] (0,0) -- (0.3,0);}~\textit{graspable},
    \protect\raisebox{2pt}{\protect\tikz \protect\draw[tcontain,line width=2] (0,0) -- (0.3,0);}~\textit{contain},
    \protect\raisebox{1pt}{\protect\tikz \protect\draw[pattern color=black] (0,0) rectangle (0.3,0.1);}~\textit{arm}.}
    \label{fig:qual_ho3d_varying_scale}
\end{figure}

Fig.~\ref{fig:qual_ho3d_varying_scale} compares the predictions of M2F-AFF and ACANet with the annotation. When the objects are scaled down, ACANet predicts false positives for the class \textit{arm} on the image padding. Both models predicts false positive for the \textit{graspable} class in case of clutter (top image). M2F-AFF has higher false negatives for the \textit{arm} class than ACANet with the original image resolution (last row in both images). 

\end{document}